\documentclass[10pt,twocolumn,letterpaper,table,hyphens]{article}

\usepackage{booktabs} %
\usepackage{colortbl}
\usepackage{tabularx}
\usepackage{multirow}
\usepackage[table,x11names,dvipsnames,table]{xcolor}
\usepackage{wrapfig}

\definecolor{rowblue}{RGB}{220,230,240}
\definecolor{myorchid}{RGB}{150,10,30}
\definecolor{myblue}{RGB}{10,30,250}
\definecolor{mygreen}{RGB}{10,190,10}

\newcommand{\supp}[1]{#1}

\newcommand{\camready}[1]{#1}

\newcommand{\myparagraph}[1]{\vspace{.2cm} \noindent \textbf{#1} \quad}
\newcommand{\etal}{\textit{et al.}}

\renewcommand{\thefootnote}{\fnsymbol{footnote}}
\newcommand{\nummodels}{11\xspace}

\newcommand{\dsetname}{ForenSynths\xspace}

\newcommand{\fig}[1]{Figure~\ref{#1}}
\newcommand{\tbl}[1]{Table~\ref{#1}}

\usepackage{enumitem}

\makeatletter
\renewcommand{\paragraph}{%
  \@startsection{paragraph}{4}%
  {\z@}{0.8ex \@plus 1ex \@minus .2ex}{-1em}%
  {\normalfont\normalsize\bfseries}%
}
\makeatother

\usepackage{cvpr}
\usepackage{times}
\usepackage{epsfig}
\usepackage{graphicx}
\usepackage{amsmath}
\usepackage{amssymb}

\usepackage{pifont}

\usepackage{subfiles}
\usepackage{color}
\usepackage{enumitem}
\usepackage{dsfont}
\usepackage{booktabs}
\usepackage{float}
\usepackage{capt-of,etoolbox}
\usepackage[font={small}]{caption}
\usepackage{subcaption}
\usepackage{multirow}
\usepackage{arydshln}
\usepackage{amsmath}
\usepackage{bbding}

\definecolor{MyDarkBlue}{rgb}{0.08,0.02,0.5}
\definecolor{MyLightBlue}{rgb}{0,0.08,0.8}
\definecolor{MyDarkGreen}{rgb}{0.02,0.50,0.02}
\definecolor{MyRed}{rgb}{1.,0.,0.}
\definecolor{MyDarkRed}{rgb}{0.7,0.02,0.02}
\definecolor{MyDarkOrange}{rgb}{0.40,0.2,0.02}
\definecolor{MyDarkMagenta}{rgb}{0.337,0,0.827}
\definecolor{MyDarkGray}{rgb}{0.5,0.5,0.5}
\newcommand{\gray}[1]{\textcolor{MyDarkGray}{#1}}
\newcommand{\red}[1]{\textcolor{MyRed}{#1}}
\newcommand{\style}[1]{#1}

\usepackage[pagebackref=true,breaklinks=true,letterpaper=true,colorlinks,bookmarks=false]{hyperref}

\cvprfinalcopy %

\def\cvprPaperID{36} %

\ifcvprfinal\fi
\begin{document}

\title{CNN-generated images are surprisingly easy to spot... for now}

\makeatletter
\def\and{%
  \end{tabular}%
  \hskip 0.995em \@plus.17fil\relax
  \begin{tabular}[t]{c}}
\makeatother

\author{Sheng-Yu Wang\textsuperscript{1}
\and
Oliver Wang\textsuperscript{2}
\and
Richard Zhang\textsuperscript{2}
\and
Andrew Owens\textsuperscript{1,3}
\and
Alexei A. Efros\textsuperscript{1}
\and
\\\and
UC Berkeley\textsuperscript{1}
\and
Adobe Research\textsuperscript{2}
\and
University of Michigan\textsuperscript{3}
}

\twocolumn[{%
\renewcommand\twocolumn[1][]{#1}%
\maketitle

\begin{center}
    
    \centering
    {\scriptsize
    \def\teaserwid{0.085\linewidth}
    \begin{tabular}{c@{\hspace{.5mm}}*{11}{c@{\hspace{.5mm}}}}
        \rotatebox[origin=c]{90}{synthetic}&
         \raisebox{-0.45\height}{\includegraphics[width=\teaserwid]{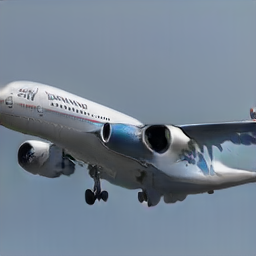}}&  
         \raisebox{-0.45\height}{\includegraphics[width=\teaserwid]{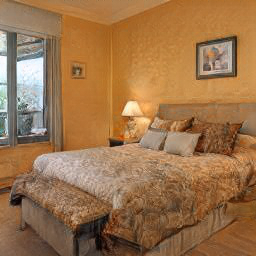}}&
         \raisebox{-0.45\height}{\includegraphics[width=\teaserwid]{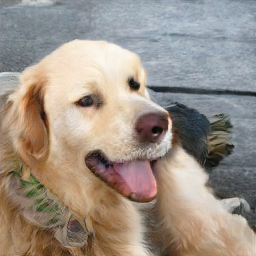}}&
         \raisebox{-0.45\height}{\includegraphics[width=\teaserwid]{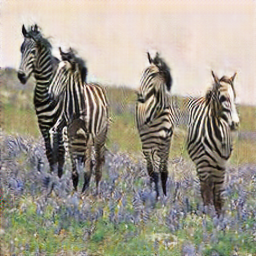}}&
         \raisebox{-0.45\height}{\includegraphics[width=\teaserwid]{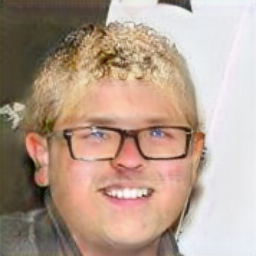}}&
         \raisebox{-0.45\height}{\includegraphics[width=\teaserwid]{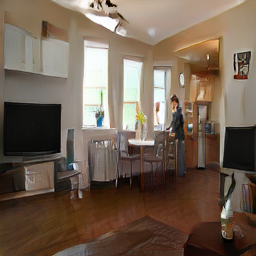}}&
         \raisebox{-0.45\height}{\includegraphics[width=\teaserwid,height=\teaserwid]{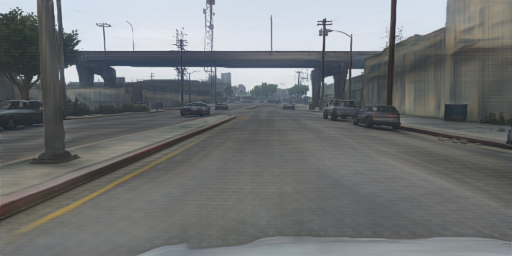}}&
         \raisebox{-0.45\height}{\includegraphics[width=\teaserwid,height=\teaserwid]{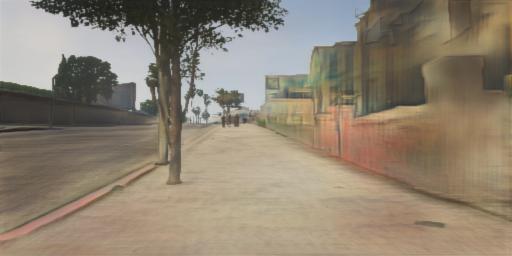}}&
         \raisebox{-0.45\height}{\includegraphics[width=\teaserwid,height=\teaserwid]{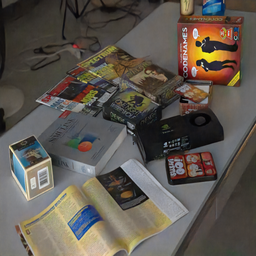}}&
         \raisebox{-0.45\height}{\includegraphics[width=\teaserwid,height=\teaserwid]{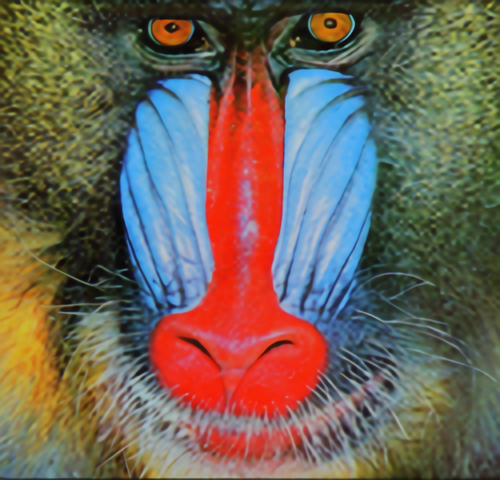}}&
         \raisebox{-0.45\height}{\includegraphics[width=\teaserwid,height=\teaserwid]{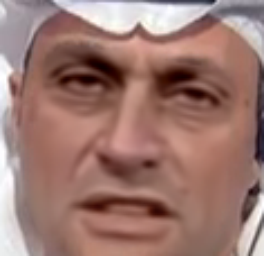}}\\[.65cm]

        \rotatebox[origin=c]{90}{real}&
         \raisebox{-0.45\height}{\includegraphics[width=\teaserwid]{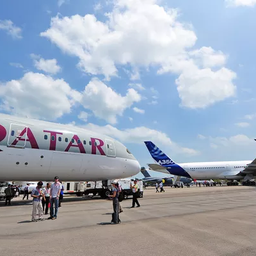}}&  
         \raisebox{-0.45\height}{\includegraphics[width=\teaserwid]{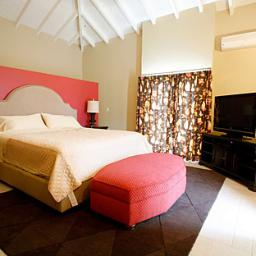}}&
         \raisebox{-0.45\height}{\includegraphics[width=\teaserwid]{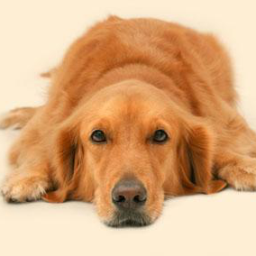}}&
         \raisebox{-0.45\height}{\includegraphics[width=\teaserwid]{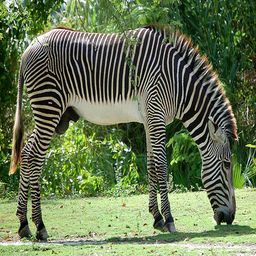}}&
         \raisebox{-0.45\height}{\includegraphics[width=\teaserwid]{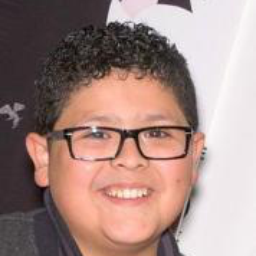}}&
         \raisebox{-0.45\height}{\includegraphics[width=\teaserwid]{figures/teaser/gau-real.png}}&
         \raisebox{-0.45\height}{\includegraphics[width=\teaserwid,height=\teaserwid]{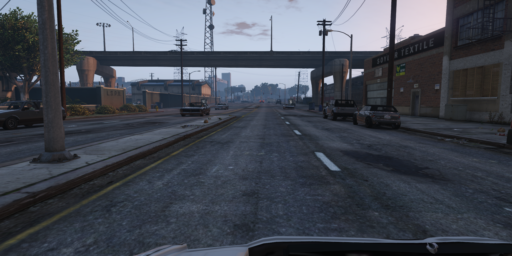}}&
         \raisebox{-0.45\height}{\includegraphics[width=\teaserwid,height=\teaserwid]{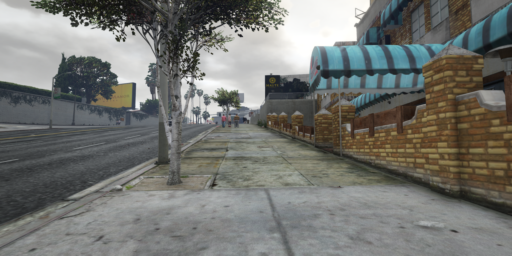}}&
         \raisebox{-0.45\height}{\includegraphics[width=\teaserwid,height=\teaserwid]{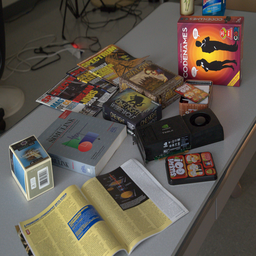}}&
         \raisebox{-0.45\height}{\includegraphics[width=\teaserwid,height=\teaserwid]{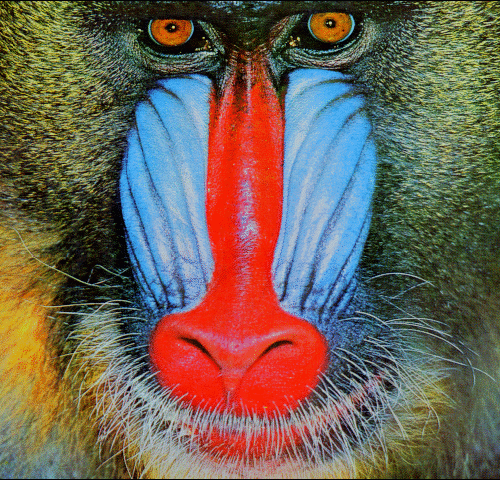}} &
         \raisebox{-0.45\height}{\includegraphics[width=\teaserwid,height=\teaserwid]{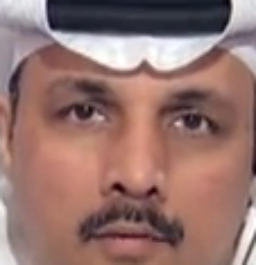}}
        \vspace{.05in}
         \\
         &
         ProGAN~\cite{karras2018progressive} &  
         StyleGAN~\cite{karras2019style} &  
         BigGAN~\cite{brock2018large} &  
         CycleGAN~\cite{zhu2017unpaired} &
         StarGAN~\cite{choi2018stargan} &  
         GauGAN~\cite{park2019SPADE} &  
         CRN~\cite{chen2017photographic} &
         IMLE~\cite{li2019diverse} &
         SITD~\cite{chen2018learning} &
         Super-res.~\cite{dai2019second} &
         Deepfakes~\cite{rossler2019faceforensics++} \\
    \end{tabular}}
    \vspace{-.1in}
    \captionof{figure}{{\bf Are CNN-generated images hard to distinguish from real images?} We show that a classifier trained to detect images generated by only one CNN (ProGAN, far left) can detect those generated by many other models (remaining columns). Our code and models are available at \url{https://peterwang512.github.io/CNNDetection/}.
    \label{fig:teaser}
    }

\end{center}
}] 
\maketitle

\begin{abstract}
\style{In this work we ask whether it is possible to create a ``universal'' detector for telling apart real images from these generated by a CNN, regardless of architecture or dataset used.} 
To test this, we collect a dataset consisting of fake images generated by \nummodels different CNN-based image generator models, chosen to span the space of commonly used architectures today
\style{(ProGAN, StyleGAN, BigGAN, CycleGAN, StarGAN, GauGAN, DeepFakes, cascaded refinement networks, implicit maximum likelihood estimation, second-order attention super-resolution, seeing-in-the-dark)}.
We demonstrate that, with careful pre- and post-processing and data augmentation, a standard image classifier trained on {\em only one} specific CNN generator \style{(ProGAN)} is able to generalize surprisingly well to unseen architectures, datasets, and training methods 
\style{(including the just released StyleGAN2~\cite{karras2019analyzing}).}
Our findings suggest the intriguing possibility that today's CNN-generated images share some common systematic flaws, preventing them from achieving realistic image synthesis. 
\end{abstract} %
\section{Introduction}

Recent rapid advances in deep image synthesis techniques, such as Generative Adversarial Networks (GANs),
have generated a huge amount of public interest and concern, as people worry that we are entering a world where it will be impossible to tell which images are real and which are fake~\cite{farid2016photo}.  
This issue has
started to play a significant role in global politics; in one case a video of the president of Gabon that was claimed by opposition to be fake was one factor leading to a failed coup d'etat\footnote{\url{https://www.motherjones.com/politics/2019/03/deepfake-gabon-ali-bongo/}}.
Much of this concern has been directed at specific manipulation techniques, such as ``deepfake"-style face replacement~\cite{deepfakes}, and photo-realistic synthetic humans~\cite{karras2019style}.
However, these methods represent only two \emph{instances} of a broader set of techniques: image synthesis via convolutional neural networks (CNNs). 
Our goal in this work is to find a general image forensics approach for detecting CNN-generated imagery.

Detecting whether an image was generated by a specific synthesis technique is relatively straightforward ---
just train a classifier on a dataset consisting of real images and images synthesized by the technique in question.
However, such an approach will likely be tied to the dataset used in image generation (\eg faces), and, due to dataset bias~\cite{torralba2011unbiased}, might not generalize when tested on new data (\eg cars).
Even worse, the technique-specific detector is likely to soon become ineffective as generation methods evolve and the technique it was trained on becomes obsolete.

It is natural, therefore, to ask whether today's CNN-generated images contain common artifacts, \eg, some kind of detectable CNN \emph{fingerprints}, that would allow a classifier to generalize to an entire family of generation methods, rather than a single one.
Unfortunately, prior work has reported generalization to be a significant problem for image forensics approaches.
For example, several recent works~\cite{zhang2019detecting,cozzolino2018forensictransfer,wang2019fakespotter} observe that that classifiers trained on images produced by one GAN architecture perform poorly when tested on others, and in many cases they also fail to generalize when only the dataset (and not the architecture or task) is changed~\cite{zhang2019detecting}.
This makes sense, as image generation methods are highly varied: they use different datasets, network architectures, loss functions, and image pre-processing.

In this paper, we show that, contrary to this current understanding, classifiers trained to detect CNN-generated images \emph{can} exhibit a surprising amount of generalization ability across datasets, architectures, and tasks.
We follow conventions and train our classifiers in a straightforward manner, by generating a large number of fake images using a {\em single} CNN model (we use ProGAN, a high-performing unconditional GAN model~\cite{karras2018progressive}), and train a binary classifier to detect fakes, using the model's real training images as negative examples.

To evaluate our model, we create a new dataset of CNN-generated images, the {\em \dsetname} dataset, consisting of synthesized images from \nummodels models, that range from from unconditional image generation methods, such as StyleGAN~\cite{karras2019style}, to super-resolution methods~\cite{dai2019second}, and deepfakes~\cite{rossler2019faceforensics++}. %
Each model is trained on a different image dataset appropriate for its specific task. 
\style{We have also continued evaluating our detector on models that were released after our paper was originally written, finding that it works out-of-the-box on the very recent unconditional GAN, StyleGAN2~\cite{karras2019analyzing}.}

Underneath the apparent simplicity of this approach, we have found that there are a number of subtle challenges which we study through a set of experiments and a new dataset of trained image generation models. 
We find that data augmentation, in the form of common image post-processing operations, is critical for generalization, even when the target images are not post-processed themselves. 
We also find that diversity of training images matters; large datasets sampled from CNN synthesis methods lead to classifiers that outperform those trained on smaller datasets, to a point. 
Finally, it is critical to examine the effect of post-processing on the model's generalization ability which often occur downstream of image creation (\eg, during storage and distribution).
We show that when the correct steps are taken, classifiers are indeed robust to common operations such as JPEG compression, blurring, and resizing.

In summary, our main contributions are: 1) we show that forensics models trained on CNN-generated images exhibit a surprising amount of generalization to other CNN synthesis methods; 2) we propose a new dataset and evaluation metric for detecting CNN-generated images; 3) we experimentally analyze the factors that account for cross-model generalization.

\section{Related work}

\paragraph{Detecting CNN-based Manipulations} Several recent works have addressed the problem of detecting images generated by CNNs. R{\"o}ssler \etal~\cite{rossler2019faceforensics++} evaluated methods for detecting face manipulation techniques, including CNN-based face and mouth replacement methods. While they showed that simple classifiers could detect fakes generated {\em by the same} model, they did not study generalization between models or datasets. Marra \etal~\cite{marra2018detection} likewise showed that simple classifiers can detect images created by an image translation network~\cite{isola2017image}, but did not consider cross-model transfer. 

Recently, Cozzolino~\etal~\cite{cozzolino2018forensictransfer} found that forensics classifiers transferred poorly between models, often obtaining near-chance performance. 
They propose a new representation learning method, based on autoencoders, to improve transfer performance in zero- and low-shot training regimes for a variety of generation methods. 
While their ultimate goal is similar to ours, they take an orthogonal approach. They focus on new learning methods for improving transfer learning, and apply them to a diverse assortment of models (including both CNN and non-CNN).
In contrast, we empirically study the performance of simple ``baseline" classifiers under different training and testing conditions for CNN-based image generation. 
Zhang~\etal~\cite{zhang2019detecting} finds that classifiers generalize poorly between GAN models. 
They propose a method called AutoGAN for generating images that contain the upsampling artifacts common in GAN architectures, and test it on two types of GANs. 
Other work has proposed to detect GAN images using hand-crafted co-occurrence features~\cite{nataraj2019detecting}, or by anomaly detection models built on pretrained face detectors~\cite{wang2019fakespotter}. Researchers have also proposed methods for identifying which, of several, known GANs generated a given image~\cite{marra2019gans,yu2019attributing}.

\paragraph{Image forensics} Researchers have proposed a variety of methods for detecting more traditional manipulation techniques, such as those made by image editing tools. Early work focused on hand-crafted cues~\cite{farid2016photo} such as compression artifacts~\cite{agarwal2017photo}, resampling~\cite{popescu2005exposing}, or physical scene constraints~\cite{o2012exposing}. More recently, researchers have applied learning-based methods to these problems~\cite{zhou2016learning,huh2018fighting,cozzolino2015splicebuster,rao2016deep,wang2019detecting}. This line of work has found, like us, that simple, supervised classifiers are often effective at detecting manipulations~\cite{zhou2016learning,wang2019detecting}.

\paragraph{Artifacts from CNN-based Generators} Researchers have shown, recently, that common CNN designs contain artifacts that reduce their representational power. Much of this work has focused on the way networks perform upampling and downsampling. A well-known example of such an artifact is the checkerboard artifact produced by deconvolutional layers~\cite{odena2016deconvolution}. Azulay and Weiss~\cite{azulay2018deep} showed convolutional networks ignore the classical sampling theorem and that strided convolutions therefore reduce translation invariance, and Zhang~\cite{zhang2019making} improved translation invariance by reducing aliasing in these layers. Very recently, Bau \etal~\cite{bau2019seeing} suggested that GANs have limited generation capacity, and analyzed the image structures that a pretrained GAN is unable to produce.

\section{A dataset of CNN-based generation models}

To study the transferability of classifiers trained to detect CNN-generated images, we collected a dataset of images created from a variety of CNN models. %

\subsection{Generation models} 
\label{sec:dataset_detail}
Our dataset contains \nummodels synthesis models. 
We chose methods that span a variety of CNN architectures, datasets, and losses. 
All of these models have an upsampling-convolutional structure (\ie they generate images by a series convolution and upsampling operations) since this is by far the most common design for generative CNNs.
Examples of their synthesized images can be found in \fig{fig:teaser}. The statistics of each dataset are listed in Table~\ref{tab:datastats}. \supp{Details of the data collection process are provided in Appendix~\ref{sec:data_appendix}.}

\paragraph{GANs} We include three state-of-the-art unconditional GANs: ProGAN~\cite{karras2018progressive}, StyleGAN~\cite{karras2019style}, BigGAN~\cite{brock2018large}, trained on either the LSUN~\cite{yu2015lsun} or ImageNet~\cite{russakovsky2015imagenet} datasets. 
The network structures and training procedures for these models contain significant differences. 
ProGAN and StyleGAN train a different network for each category; StyleGAN injects large, per-pixel noise into the model to introduce high frequency detail. BigGAN has a monolithic, class-conditional structure, is trained on very large batch sizes, and uses self-attention layers~\cite{zhang2018self,wang2018non}.

We also include three conditional GANs: the state-of-the-art image-to-image translation method GauGAN~\cite{park2019SPADE}, and the popular unpaired image-to-image translation methods CycleGAN~\cite{zhu2017unpaired} and StarGAN~\cite{choi2018stargan}. 

\paragraph{Perceptual loss} We consider models that directly optimize a perceptual loss~\cite{johnson2016perceptual}, with no adversarial training. This includes Cascaded Refinement Networks (CRN)~\cite{chen2017photographic}, which synthesizes images in a  coarse-to-fine manner, and the recent Implicit Maximum Likelihood Estimation (IMLE) conditional image translation model~\cite{li2019diverse}. 

\paragraph{Low-level vision} We include the Seeing In The Dark (SITD) model~\cite{chen2018learning}, which approximates long-exposure photography under low light conditions from short-exposure raw camera input using a high-resolution fully convolutional network. 
We also use a state-of-the-art super-resolution model, the Second Order Attention Network (SAN)~\cite{dai2019second}.%

\paragraph{Deep fakes} We also evaluate our model on the face replacement images provided in the FaceForensics++ benchmark of R{\"o}ssler \etal~\cite{rossler2019faceforensics++}, which used the publicly available {\em faceswap} tool~\cite{deepfakes2019faceswap}. While ``deepfake" is often used as a general term, we take inspiration from the convention in~\cite{rossler2019faceforensics++} and refer to this specific model as {\em DeepFake}.
This model uses an autoencoder to generate faces, and images undergo extensive post-processing steps, including Poisson image blending~\cite{perez2003poisson} with real content. \camready{We note that our main goal is to detect images directly output by CNN decoders, while DeepFake serves as an out-of-distribution test case. } Following~\cite{rossler2019faceforensics++}, we use cropped faces. %

\begin{table}[t]
    \centering 
    \resizebox{1.\linewidth}{!}{
    {\def\arraystretch{1.2}
     \begin{tabular}{@{}lllr@{}}
    \toprule Family & Method & Image Source  & \# Images \\
    \midrule
    \multirow{3}{*}{\shortstack[l]{Unconditional\\GAN}} & ProGAN~\cite{karras2018progressive} & LSUN & 8.0k  \\
    & StyleGAN~\cite{karras2019style} & LSUN & 12.0k  \\
    & BigGAN~\cite{brock2018large} & ImageNet & 4.0k  \\ \cdashline{1-4}
    \multirow{3}{*}{\shortstack[l]{Conditional\\GAN}} & CycleGAN~\cite{zhu2017unpaired} & Style/object transfer & 2.6k \\
    & StarGAN~\cite{choi2018stargan} & CelebA & 4.0k  \\
    & GauGAN~\cite{park2019SPADE} & COCO & 10.0k \\ \cdashline{1-4}
    \multirow{2}{*}{\shortstack[l]{Perceptual\\loss}} & CRN~\cite{chen2017photographic} & GTA & 12.8k  \\
    & IMLE~\cite{li2019diverse} & GTA & 12.8k \\  \cdashline{1-4}
    \multirow{2}{*}{\shortstack[l]{Low-level\\vision}} & SITD~\cite{chen2018learning} & Raw camera & 360  \\
    & SAN~\cite{dai2019second} & Standard SR benchmark & 440  \\  \cdashline{1-4}
    Deepfake & FaceForensics++~\cite{rossler2019faceforensics++} & Videos of faces & 5.4k \\
    \bottomrule
    \end{tabular}}
    }
    \caption{{\bf Generation models.} We evaluate forensic classifiers on a variety of CNN-based image generation methods.}
    \label{tab:datastats}
    \vspace{-5mm}
\end{table} 
\subsection{Generating fake images}

We collect images from the models, taking care to match the  pre-processing operations performed by each (\eg resizing and cropping). 
For each dataset, we collect fake images by generating them from the model without applying additional post-processing (or we download the officially released generated images if they are available).
We collect an equal number of real images from each method's training set.
To make the distribution of the real and fake images as close as possible, real images are pre-processed according to the pipeline prescribed by each method.

Since $256\times256$ resolution is the most commonly shared output size among most off-the-shelf image synthesis models (\eg, CycleGAN, StarGAN, ProGAN LSUN, GauGAN COCO, IMLE, \etc), we used this resolution for our dataset.
For models that produce images at lower resolutions, (\eg, DeepFake), we rescale the images using bilinear interpolation to $256$ on the shorter side with the same aspect ratio, and for models that produce images at higher resolution (\eg, ProGAN, StyleGAN, SAN, SITD), we keep the images at the same resolution. 
Despite these cases being slightly different from our training scheme, we observe that our model is still able to detect fake images under these categories. For all datasets, we make our real/fake prediction from $224\times224$ crops \camready{(random-crop at training time and center-crop at testing time)}.

\newcolumntype{N}{>{\centering\arraybackslash\hsize=.09\hsize}X}

\begin{table*}[h]
  {\small
    \centering
    \resizebox{1.\linewidth}{!}{
    \begin{tabular}{ccccccc ccccccccccc c}
    \toprule

    \multirow{3}{*}{Family} & \multirow{3}{*}{Name} & \multicolumn{5}{c}{Training settings} & \multicolumn{11}{c}{Individual test generators} & Total \\
    \cmidrule(lr){3-7} \cmidrule(lr){8-18} \cmidrule(lr){19-19}

    & & \multirow{2}{*}{Train} & \multirow{2}{*}{Input} & \multirow{2}{*}{\shortstack[c]{No.\\Class}} & \multicolumn{2}{c}{Augments} & \multirow{2}{*}{\shortstack[c]{Pro-\\GAN}} & \multirow{2}{*}{\shortstack[c]{Style-\\GAN}} & \multirow{2}{*}{\shortstack[c]{Big-\\GAN}} & \multirow{2}{*}{\shortstack[c]{Cycle-\\GAN}} & \multirow{2}{*}{\shortstack[c]{Star-\\GAN}} & \multirow{2}{*}{\shortstack[c]{Gau-\\GAN}} & \multirow{2}{*}{CRN} & \multirow{2}{*}{IMLE} & \multirow{2}{*}{SITD} & \multirow{2}{*}{SAN} & \multirow{2}{*}{\shortstack[c]{Deep-\\Fake}} &
     \multirow{2}{*}{\bf mAP}\\ 
    \cmidrule(lr){6-7}
    & & & & & Blur & JPEG & \\
    \midrule

\multirow{4}{*}{\shortstack[c]{Zhang \\et al.\\~\cite{zhang2019detecting}}} & Cyc-Im & CycleGAN & RGB & -- & & & 84.3 & 65.7 & 55.1 & \gray{100.} & 99.2 & 79.9 & 74.5 & 90.6 & 67.8 & 82.9 & 53.2 & 77.6 \\
& Cyc-Spec & CycleGAN & Spec & -- & & & 51.4 & 52.7 & 79.6 & \gray{100.} & {\bf 100.} & 70.8 & 64.7 & 71.3 & 92.2 & 78.5 & 44.5 & 73.2 \\ \cdashline{2-19}
& Auto-Im & AutoGAN & RGB & -- & & & 73.8 & 60.1 & 46.1 & 99.9 & {\bf 100.} & 49.0 & 82.5 & 71.0 & 80.1 & 86.7 & 80.8 & 75.5 \\
& Auto-Spec & AutoGAN & Spec & -- & &  & 75.6 & 68.6 & 84.9 & \bf{100.} & {\bf 100.} & 61.0 & 80.8 & 75.3 & 89.9 & 66.1 & 39.0 & 76.5 \\
\midrule
\multirow{9}{*}{Ours} & 2-class & ProGAN & RGB & 2 & \checkmark & \checkmark & \gray{98.8} & 78.3 & 66.4 & 88.7 & 87.3 & 87.4 & 94.0 & 97.3 & 85.2 & 52.9 & 58.1 & 81.3 \\
& 4-class & ProGAN & RGB & 4 & \checkmark & \checkmark & \gray{99.8} & 87.0 & 74.0 & 93.2 & 92.3 & 94.1 & 95.8 & 97.5 & 87.8 & 58.5 & 59.6 & 85.4 \\
& 8-class & ProGAN & RGB & 8 & \checkmark & \checkmark & \gray{99.9} & 94.2 & 78.9 & 94.3 & 91.9 & 95.4 & 98.9 & 99.4 & 91.2 & 58.6 & 63.8 & 87.9 \\
& 16-class & ProGAN & RGB & 16 & \checkmark & \checkmark & \gray{100.} & 98.2 & 87.7 & 96.4 & 95.5 & {\bf 98.1} & {\bf 99.0} & {\bf 99.7} & 95.3 & 63.1 & 71.9 & 91.4 \\

\cdashline{2-19}
& No aug & ProGAN & RGB & 20 &  &  & \gray{100.} & 96.3 & 72.2 & 84.0 & {\bf 100.} & 67.0 & 93.5 & 90.3 & 96.2 & {\bf 93.6} & {\bf 98.2} & 90.1 \\
& Blur only & ProGAN & RGB & 20 & \checkmark &  & \gray{100.} & 99.0 & 82.5 & 90.1 & {\bf 100.} & 74.7 & 66.6 & 66.7 & {\bf 99.6} & 53.7 & 95.1 & 84.4 \\
& JPEG only & ProGAN & RGB & 20 &  & \checkmark & \gray{100.} & 99.0 & 87.8 & 93.2 & 91.8 & 97.5 & {\bf 99.0} & 99.5 & 88.7 & 78.1 & 88.1 & {\bf 93.0} \\ 

& Blur+JPEG (0.5) & ProGAN & RGB & 20 & \checkmark & \checkmark & \gray{100.} & 98.5 & {\bf 88.2} & 96.8 & 95.4 & {\bf 98.1} & 98.9 & 99.5 & 92.7 & 63.9 & 66.3 & 90.8 \\
& Blur+JPEG (0.1) & ProGAN & RGB & 20 & $\dagger$ & $\dagger$ & \gray{100.} & {\bf 99.6} & 84.5 & 93.5 & 98.2 & 89.5 & 98.2 & 98.4 & 97.2 & 70.5 & 89.0 & 92.6 \\

    \bottomrule
    \end{tabular}}
    }
    \vspace{-1.5mm}
    \caption{{\bf Cross-generator generalization results.} We show the average precision (AP) of various classifiers from baseline Zhang et al.~\cite{zhang2019detecting} and ours, tested across 11 generators. Symbols $\checkmark$ and $\dagger$ mean the augmentation is applied with $50\%$ or $10\%$ probability, respectively, at training. Chance is $50\%$ and best possible performance is $100\%$. When test generators are used in training, we show those results in \gray{gray} (as they are not testing generalization). Values in black show cross-generator generalization. Amongst those, the highest value is highlighted in {\bf black}. We show ablations with respect to fewer classes in ProGAN and by removing data augmentation. We report the mean AP by averaging the AP scores over all datasets. Subsets are plotted in Figures~\ref{fig:comp_augs},~\ref{fig:comp_class},~\ref{fig:comp_methods} for comparison. %
    }
    \label{tab:mainresults}
\end{table*}

\section{Detecting CNN-synthesized images}

Are there common features or artifacts shared across diverse CNN generators? 
To understand this, we study whether it is possible to train a forensics classifier on images from {\em one} model that generalize to those of {\em many} models. 

\subsection{Training classifiers} 
While all of these models are useful for {\em evaluation}, due to limitations in dataset size, not all are well-suited to training a classifier.
We take advantage of the fact that the unconditional GAN models in our dataset can synthesize arbitrary numbers of images, and choose one specific model, ProGAN~\cite{karras2018progressive} to train the detector on. 
The decision to use a single model for training most closely resembles real world detection problems, where the diversity or number of models to generalize on are \emph{unknown} at training time. 
By selecting only a single model to train on, we are computing an \emph{upper bound} on how challenging the task is --- jointly training on multiple models would make the generalization problem easier. 
We chose ProGAN since it generates high quality images and has a simple convolutional network structure.

We then create a large-scale dataset that consists {\em solely} of ProGAN-generated images and real images. 
We use 20 models each trained on a different LSUN~\cite{yu2015lsun} object category, and generate 36K train images and 200 validation images, each with equal numbers of real and fake images for each model.
In total there are 720K images for training and 4K images for validation.
\camready{To evaluate the choice of the training dataset, we also include a model that is trained \textit{solely} on the BigGAN dataset. We also consider a model that generates training images using the deep image prior~\cite{ulyanov2018deep}, rather than a GAN. The details for these models are provided in Appendix~\ref{sec:biggan_appendix} and~\ref{sec:dip_appendix}.}

The main idea of our experiments is to train a ``real-or-fake" classifier on this ProGAN dataset, and evaluate how well the model generalizes to other CNN-synthesized images. 
For the choice of classifier, we use ResNet-50~\cite{he2016deep} pre-trained with ImageNet, and train it in a binary classification setting.
\supp{Details of the
training procedure are provided in Appendix~\ref{sec:traindetail_appendix}.}

\begin{figure*}[h]
    \centering
    \includegraphics[width=1.\linewidth]{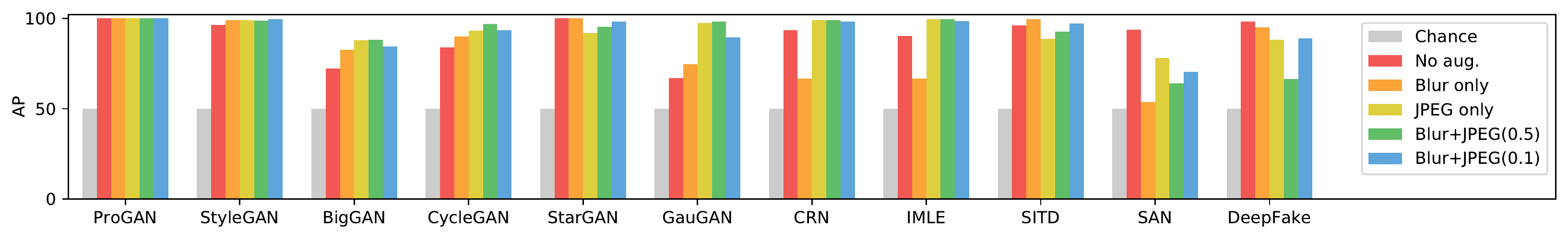}\vspace{-3.5mm}
    \caption{{\bf Effect of augmentation methods.} All detectors are trained on ProGAN, and tested on other generators (AP shown). In general, training with augmentation helps performance. Notable exceptions are super-resolution and DeepFake.}
    \label{fig:comp_augs}
\vspace*{\floatsep}
    \includegraphics[width=1.\linewidth]{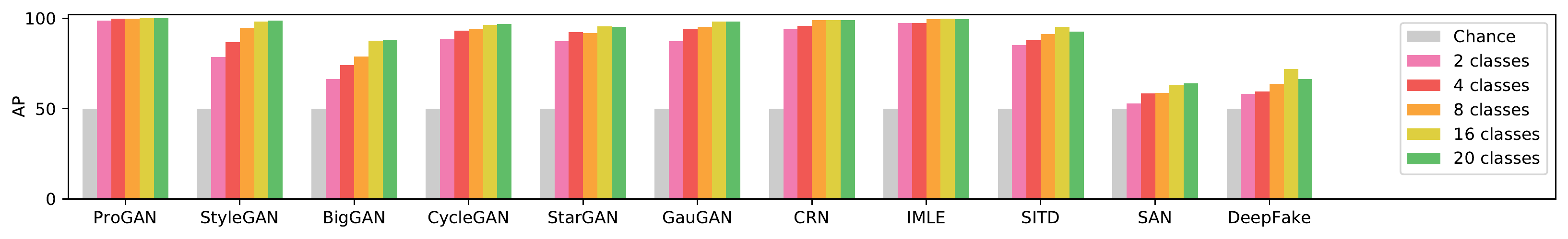}
    \caption{
    {\bf Effect of dataset diversity.} All detectors are trained on ProGAN, and tested on other generators (AP shown). Training with more classes improves performance. All runs use blur and JPEG augmentation with $50\%$ probability.}
    \label{fig:comp_class}
\vspace*{\floatsep}
    \includegraphics[width=1.\linewidth]{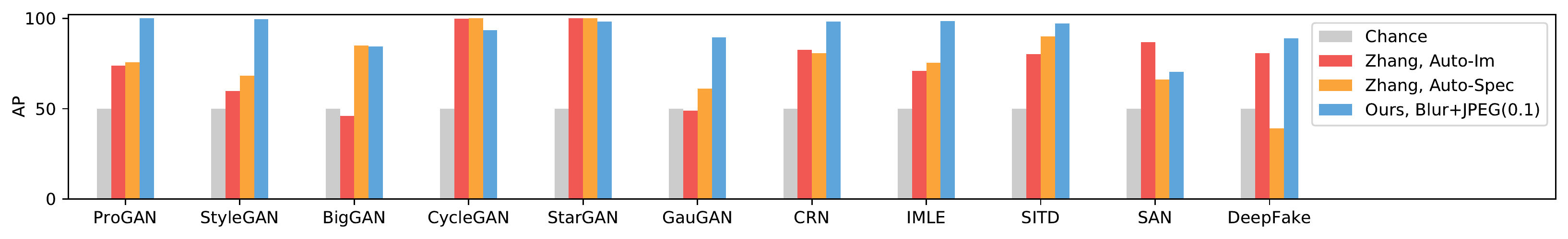}
    \caption{
    {\bf Model comparison.} Compared to Zhang~\etal~\cite{zhang2019detecting}, we observe that for the most part, our models generalize better to other architectures. Notable exceptions to this are CycleGAN (which is identical to the training architecture from ~\cite{zhang2019detecting}), StarGAN (where both methods obtain close to 100. AP), and SAN (where applying data augmentation hurts performance).
    }
    \vspace{-2mm}
    \label{fig:comp_methods}
\end{figure*}

\paragraph{Data augmentation} During training, we simulate image post-processing operations in a variety of ways. All of our models are trained with images that are randomly
left-right flipped and cropped to 224 pixels. 
We evaluate several additional augmentation variants: {\bf (1) No aug}: no augmentation applied, {\bf (2) Gaussian blur}: before cropping, with 50\% probability, images are blurred with $\sigma \sim \mbox{Uniform}[0,3]$, {\bf (3) JPEG}: with 50\% probability images are JPEG-ed by two popular libraries, OpenCV~\cite{bradski2008learning} and the Python Imaging Library (PIL), with quality $\sim \mbox{Uniform}\{30,31,\dots,100\}$, {\bf (4a) Blur+JPEG (0.5)}: the image is possibly blurred and JPEG-ed, each with 50\% probability, {\bf (4b) Blur+JPEG (0.1)}: similar to (4a), but with 10\% probability.

 \begin{figure*}[]
    \centering
    \hspace{5mm} \small{\bf Robustness to Blur} \hspace{60mm} \small{\bf Robustness to JPEG} \\
    \includegraphics[width=0.47\linewidth]{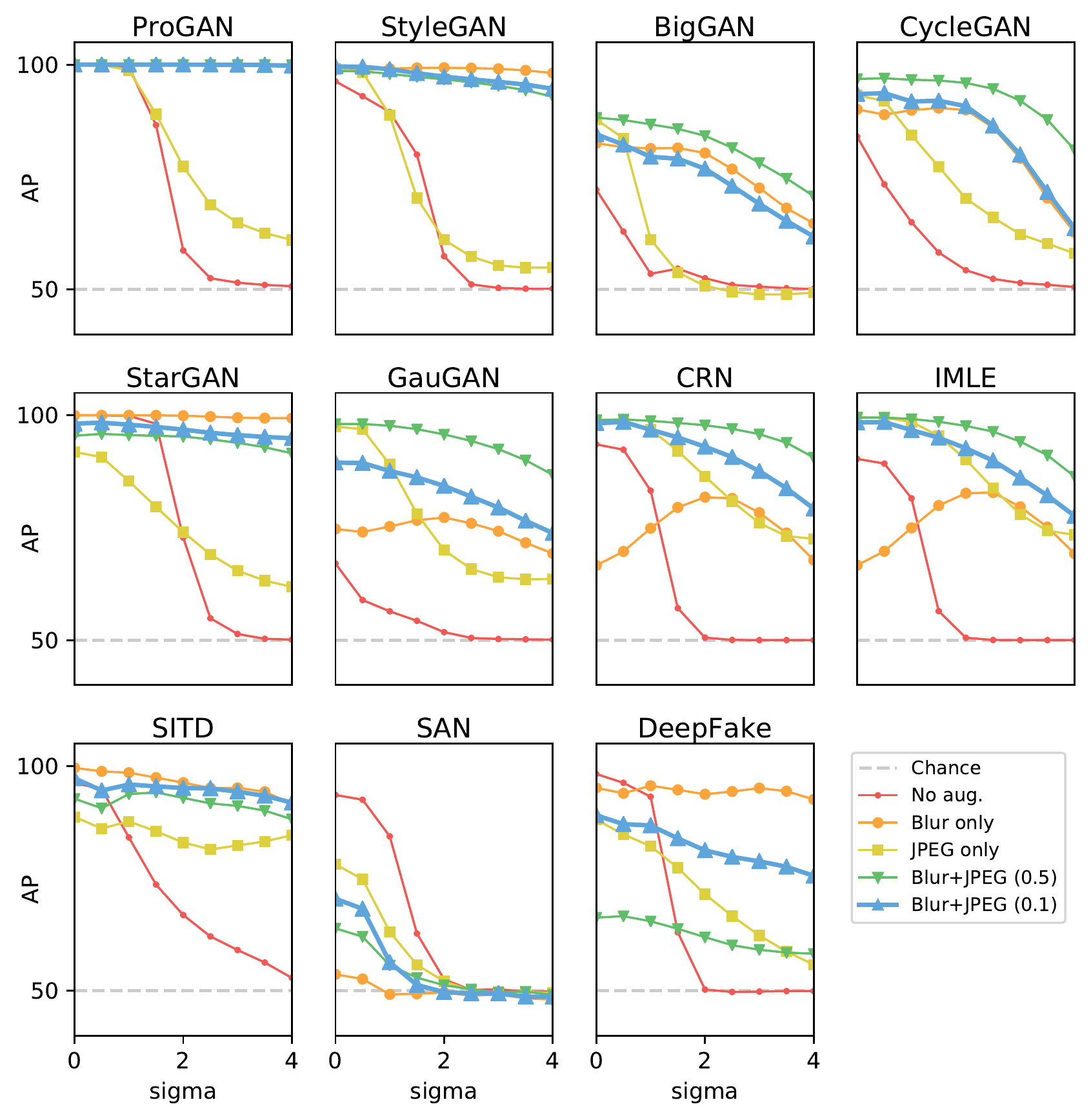} \hspace{5mm}
    \includegraphics[width=0.47\linewidth]{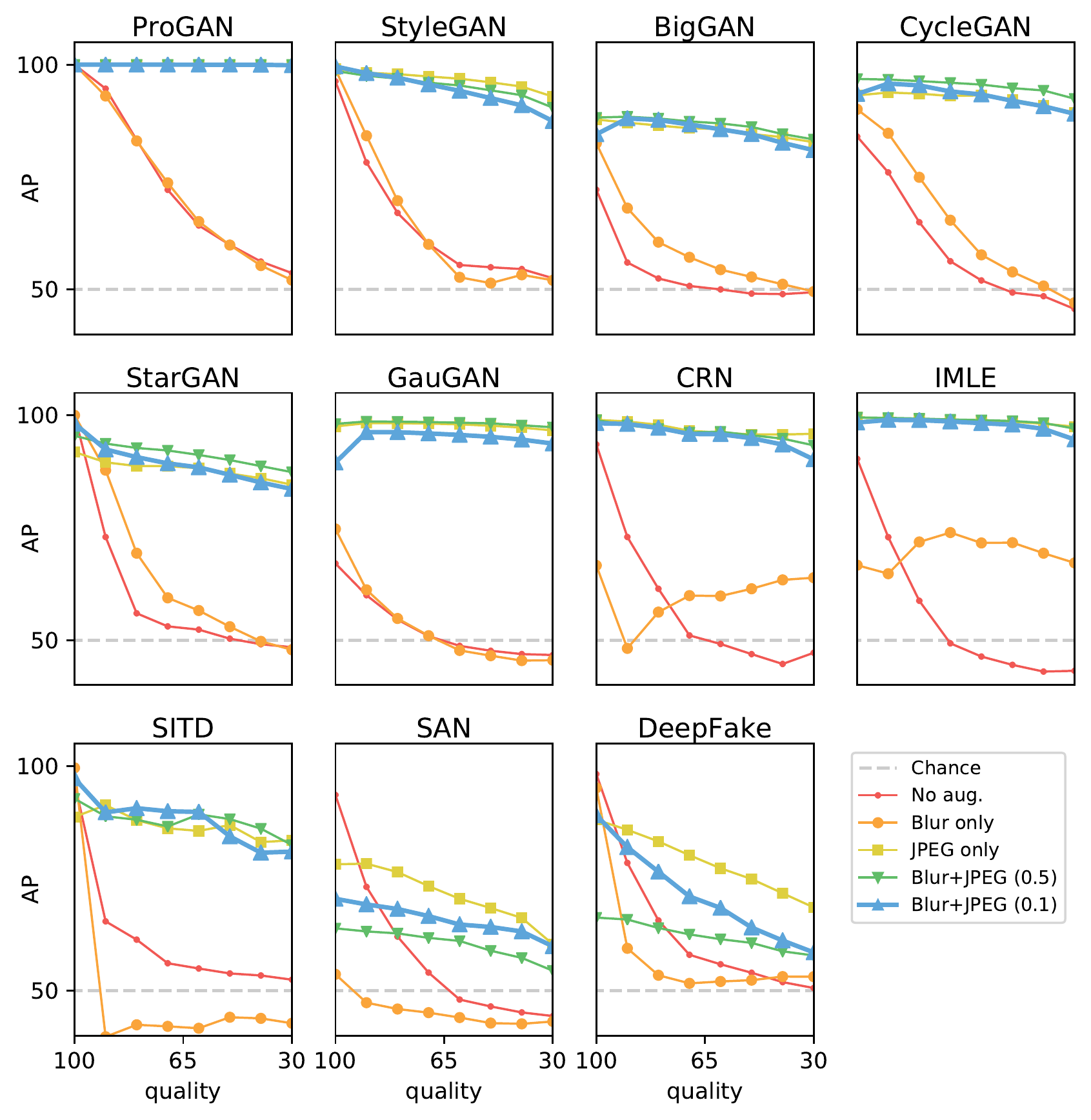}
    \caption{{\bf Robustness.} We show the effect of AP given test-time perturbation to \textbf{(left)} Gaussian blurring and \textbf{(right)} JPEG. We show classifiers trained on ProGAN, with different augmentations applied during training. Note that in all cases and both perturbations, when training without augmentation (\red{red}), performance degrades across all datasets when perturbations are added. In most cases, training with \textit{both} augmentations, performs best or near best. Notable exceptions are for super-resolution (where no augmentation is best), and DeepFake, where training \emph{only} with the perturbation used during testing, rather than both, performs best.}
    \label{fig:robustness}
\end{figure*}

\paragraph{Evaluation}
Following other recent forensics works~\cite{zhou2018learning,huh2018fighting,wang2019detecting}, we evaluate our model's performance on each dataset using average precision (AP), since it is a threshold-less, ranking-based score that is not sensitive to the base rate of the real and fake images in the dataset.
We compute this score for each dataset separately, since we expect it to be dependent on the semantic content of the photos as a whole.
To help interpret the threshold-less results, we also conduct experiments on thresholding the model's outputs and computing accuracy, under the assumption that real and fake images are equally likely to appear; \camready{the details are in Appendix~\ref{sec:twoshot}}. %
During testing, each image is center-cropped \camready{to 224$\times$224 pixels} without resizing in order to match the post-processing pipeline used by models during training.
No data augmentation is included during testing; instead, we conduct experiments on model robustness under post-processing in Section~\ref{sec:dataaug}.

\subsection{Effect of data augmentation}
\label{sec:dataaug}

In Table~\ref{tab:mainresults}, we investigate the generalization ability of training with different augmentation methods.
We find that using aggressive data augmentation (in the form of simulated post-processing) provides surprising generalization capabilities, even when such perturbations are not used at test time. 
Additionally, we observe that these models are significantly more robust to post-processing (Figure~\ref{fig:robustness}).

\paragraph{Augmentation (usually) improves generalization}
To begin, we first evaluate ProGAN-based classifier \textit{without} augmentation, shown in the ``no aug'' row. 
As in previous work~\cite{rossler2019faceforensics++}, we find that testing on held-out ProGAN images works well ($100.0$ AP). 
We then test how well it generalizes to other unconditional GANs. 
We find that it generalizes extremely well to StyleGAN, which has a similar network structure, but not as well to BigGAN.
When adding augmentations, the performance on BigGAN significantly improves, $72.2\rightarrow88.2$. On conditional models (CycleGAN, GauGAN, CRN, and IMLE), performance is similarly improved, $84.0\rightarrow96.8$, $67.0\rightarrow98.1$, $93.5\rightarrow98.9$, $90.3\rightarrow99.5$, respectively.

Interestingly, there are two models, SAN and DeepFake, where directly training on ProGAN without augmentation performs strongly ($93.6$ and $98.2$, respectively), but augmentation hurts performance.
As SAN is a super-resolution model, only high-frequency components can differentiate between real and fake images. Removing such cues at training time (\eg by blurring) would therefore be likely to reduce performance.
\camready{As explained in Section~\ref{sec:dataset_detail}, DeepFake serves as an out-of-distribution test case as images are not generated by CNN architectures alone, but surprisingly our model is able to generalize to this test case. However, it remains challenging to identify clear reasons for the performance deterioration when applying augmentations.}
Applying augmentation, but at reduced rate {\bf (Blur+JPEG (0.1))}, offers a good balance: DeepFake detection is comparable to the no-augmentation case ($89.0$), while most other datasets are significantly improved over no augmentation.

\paragraph{Augmentation improves robustness}
In many real-world scenarios, images that we would like to evaluate have undergone unknown post-processing operations, such as compression and resizing. 
We investigated whether CNN-generated images can still be detected, even after these post-processing steps.
To test this, we blurred (simulating resampling) and JPEG-compressed the real and fake images following the protocol in~\cite{wang2019detecting}, and evaluated our ability to detect them (Figure~\ref{fig:robustness}). On ProGAN (\ie the case where the test distribution matches the training), performance is $100\%$ even when applying augmentation operations, indicating that artifacts may not only be high-frequency, but exist across frequency bands. In terms of cross-generator generalization, the augmented model is most robust to post-processing operations that are included in data augmentation, agreeing with observations from ~\cite{rossler2019faceforensics++,wang2019detecting,yu2019attributing,zhang2019detecting}. However, we note that our model {\em also} gains robustness from augmentation even when testing on {\em out-of-distribution} CNN models.

\subsection{Effect of data diversity}

Next, we asked how the diversity of the real and fake images in the training set affects a classifier's generalization ability.

\paragraph{Image diversity improves performance}
To study this, we varied the number of classes in the dataset used to train our real-or-fake classifier (\fig{fig:comp_class}). 
Specifically, we trained multiple classifiers, each one on a subset of the full training dataset by excluded both real and fake images derived from a specific set of LSUN classes. \camready{For all models we use the same augmentation scheme as the \textbf{Blur+JPEG (0.5)} model.}
We found that increasing the training set diversity improves performance, but only up to a point.
When the number of classes used increases from 2 to 16, AP consistently improves, but we see diminishing returns. Minimal improvement is observed when increasing from 16 to 20 classes.
This indicates that there may be a training dataset that is ``diverse enough'' for practical generalization.

\begin{figure*}[h]
    \centering
    \def\rankwidth{.095\linewidth}
    \newcommand{\rankfig}[1]{\raisebox{-0.3\height}{\includegraphics[width=\rankwidth,height=\rankwidth]{#1}}}
    \begingroup
    \renewcommand{\arraystretch}{0.5}
    \begin{tabular}{c@{\hspace{.5mm}}c@{\hspace{0mm}}c@{\hspace{2.0mm}}c@{\hspace{0mm}}c@{\hspace{2.0mm}}c@{\hspace{0mm}}c@{\hspace{2.0mm}}c@{\hspace{0mm}}c@{\hspace{2.0mm}}*{2}{c@{\hspace{0mm}}}}
         \multirow{2}{*}{\rotatebox[origin=c]{90}{\footnotesize BigGAN}}& 
         \rankfig{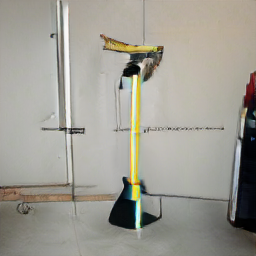}&
         \rankfig{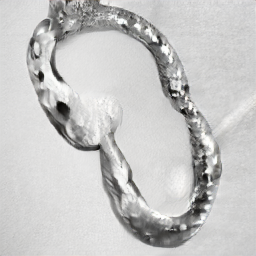}&
         
         \rankfig{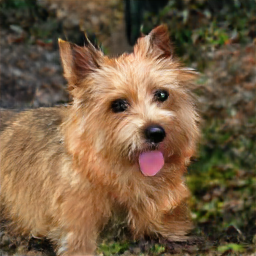}&
         \rankfig{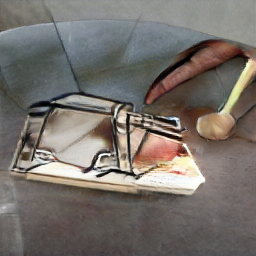}&
         
         \rankfig{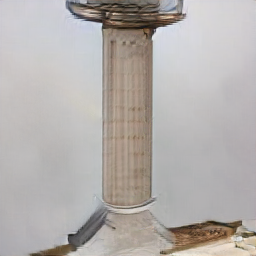}&
         \rankfig{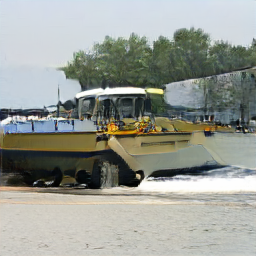}&
         
         \rankfig{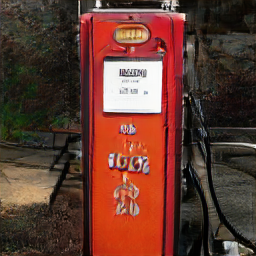}&
         \rankfig{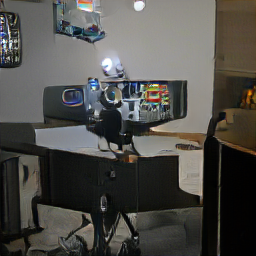}&
         
         \rankfig{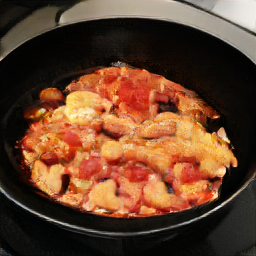}&
         \rankfig{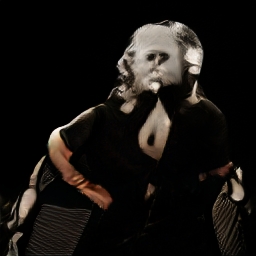}\\
         
         &
         \rankfig{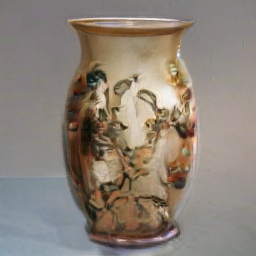}&
         \rankfig{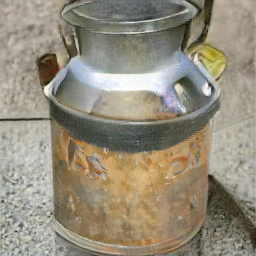}&
         
         \rankfig{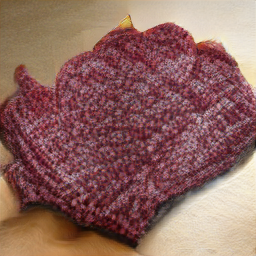}&
         \rankfig{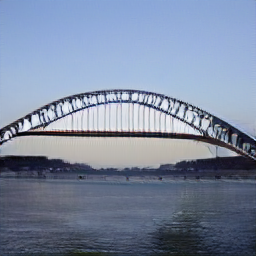}&
         
         \rankfig{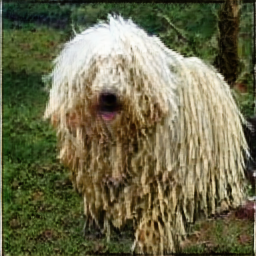}&
         \rankfig{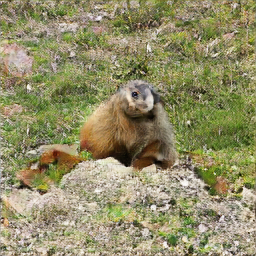}&
         
         \rankfig{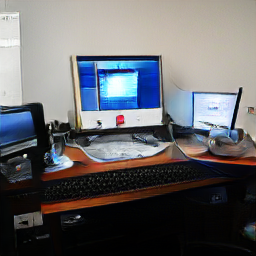}&
         \rankfig{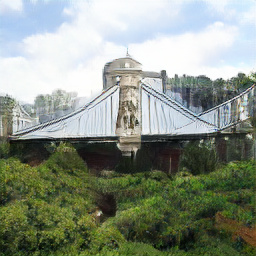}&
         
         \rankfig{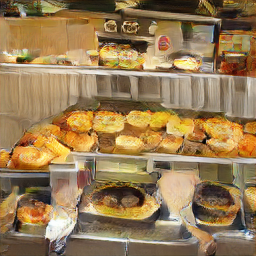}&
         \rankfig{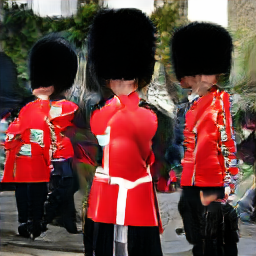}\\[.65cm]

         \multirow{2}{*}{\rotatebox[origin=c]{90}{\footnotesize StarGAN}}& \rankfig{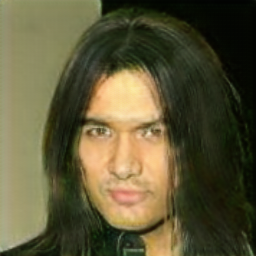}&
         \rankfig{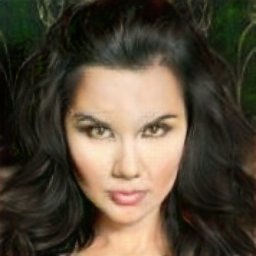}&
         
         \rankfig{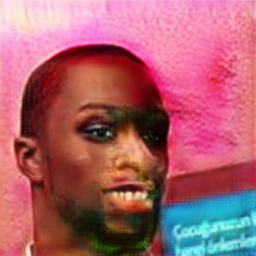}&
         \rankfig{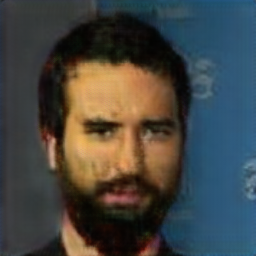}&
         
         \rankfig{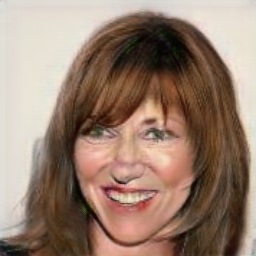}&
         \rankfig{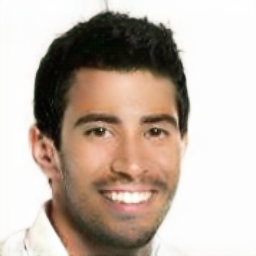}&
         
         \rankfig{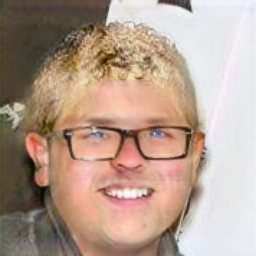}&
         \rankfig{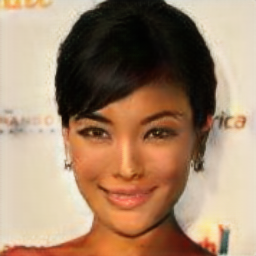}&
         
         \rankfig{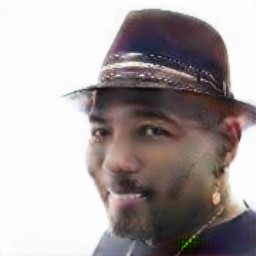}&
         \rankfig{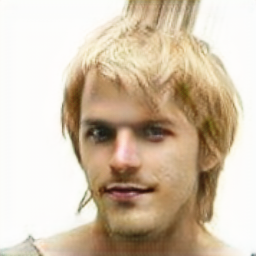}\\
         
         &
         \rankfig{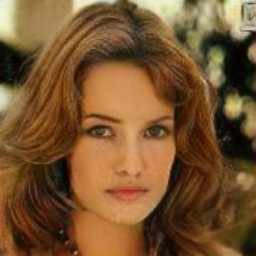}&
         \rankfig{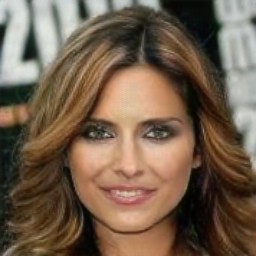}&
         
         \rankfig{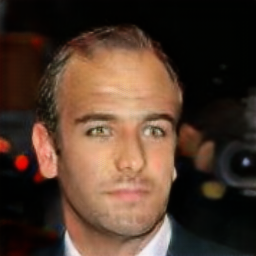}&
         \rankfig{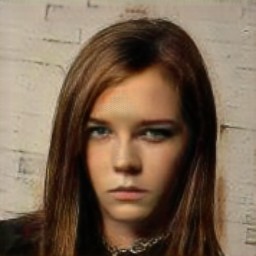}&
         
         \rankfig{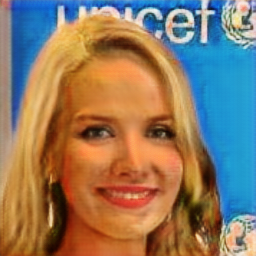}&
         \rankfig{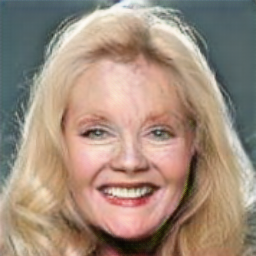}&
         
         \rankfig{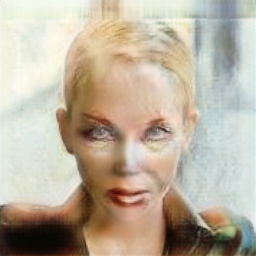}&
         \rankfig{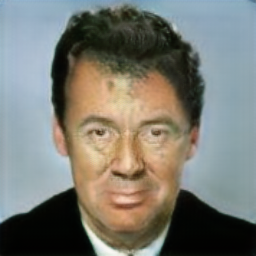}&
         
         \rankfig{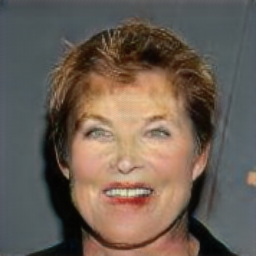}&
         \rankfig{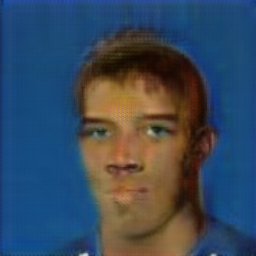}\\[.5cm]
         
         &
         \multicolumn{2}{c}{\small 0th percentile} &
         \multicolumn{2}{c}{\small 25th percentile} &
         \multicolumn{2}{c}{\small 50th percentile} &
         \multicolumn{2}{c}{\small 75th percentile} &
         \multicolumn{2}{c}{\small 100th percentile} \\
         
    \end{tabular}
    \endgroup
    \caption{{\bf Does our model's confidence correlate with visual quality?} We have found that for two models, BigGAN and StarGAN, the images on the left (considered more real) tends to look better than the images on the right (considered more fake). However, this does not seem to hold for the other models. \supp{More examples on each dataset are provided in Appendix~\ref{sec:rank_appendix}.}}
    \label{fig:ranking}
\end{figure*}

\begin{figure*}[t]
    \centering
    \includegraphics[width=1.\linewidth]{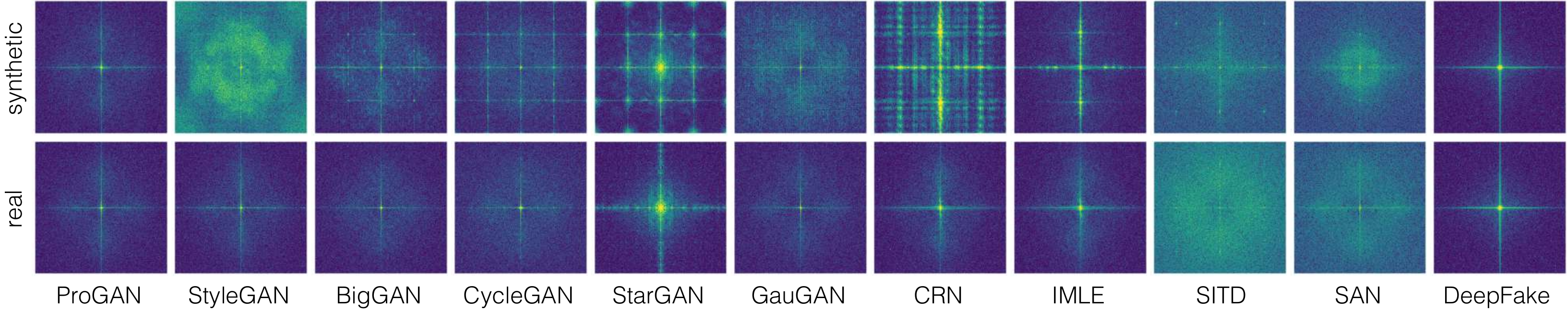}
    \caption{{\bf Frequency analysis on each dataset.} We show the average spectra of each high-pass filtered image, for both the real and fake images, similar to Zhang \etal~\cite{zhang2019detecting}. We observe periodic patterns (dots or lines) in most of the synthetic images, while BigGAN and ProGAN contains relatively few such artifacts.}
    \label{fig:spectrums}
\end{figure*}

\subsection{Comparison to other models}
Next, we asked how our generalization performance compares to other proposed forensic methods. We compare our approach to Zhang \etal~\cite{zhang2019detecting}, which is a suite of classifiers trained to detect artifacts generated by a common CNN architecture, which is shared by many image synthesis tasks such as CycleGAN and StarGAN. They introduced AutoGAN, an autoencoder based on CycleGAN's generator that simulates artifacts resembling that of CycleGAN images.

We considered four variations of pretrained models from Zhang~\etal~\cite{zhang2019detecting}, each trained from one of the two image sources (CycleGAN and AutoGAN), and one of the two image representations (images and spectrum) respectively. All four variants included JPEG and resize data augmentation during training to improve the robustness of each model.  We found that our models generalized significantly better to other architectures, except on CycleGAN (which is the model architecture used by~\cite{zhang2019detecting}), StarGAN (where both methods obtain near 100.0 AP). The comparison results are shown in Table~\ref{tab:mainresults} and Figure~\ref{fig:comp_methods}. \camready{We also include comparisons to other baseline models in Appendix~\ref{sec:baseline_appendix}.}
 
 \subsection{New CNN models}
\style{We hope that as new deep synthesis models arrive, our system will detect them out-of-the-box. One such an evaluation scenario has naturally arisen, with the recent release of StyleGAN2~\cite{karras2019analyzing}, a state-of-the-art unconditional GAN appearing in these proceedings. The StyleGAN2 model makes several changes to StyleGAN, including redesigned normalization, multi-resolution, and regularization methods. In Table~\ref{tab:style2}, we test our detector on publicly available StyleGAN2 generators. We used our \textbf{Blur+JPEG (0.1)} model and tested on the {\em LSUN} {\em car}, {\em cat}, {\em church}, and {\em horse} variants.  Despite these changes, our technique performs at 99.1$\%$ AP. These results reinforce the notion that training on today's generators can generalize well to future generators, given that they use similar underlying building blocks.}
\begin{table}[h]
    \centering
    \begin{tabular}{lccc}
    \toprule
    & ProGAN & StyleGAN & StyleGAN2 \\ \midrule
    \textbf{AP} & \gray{100.} & 99.6 & 99.1 \\

    \bottomrule
    \end{tabular}
    \caption{\small\textbf{Out-of-the box evaluation on recently released StyleGAN2~\cite{karras2019analyzing} model}. We used our \textbf{Blur+JPEG (0.1)} model and tested on StyleGAN2. We observed that our model generalizes to detecting StyleGAN2 images. Numbers for ProGAN and StyleGAN are included for comparison.}
    \label{tab:style2}
\end{table}

\style{}

\subsection{Qualitative Analysis}

To understand how the network is able to generalize to unseen CNN models, we study what possible cues the classifier might be using by visualizing its ranking on the ``fakeness'' over the synthetic dataset.
In addition, we analyze the difference between the frequency responses of both real and synthetic images across datasets. %

\myparagraph{``Fakeness'' ranking by the model} We study whether our model is learning subtle low-level features generated by CNN architectures, or high-level features such as visual quality.
Taking the similar approach as previous image realism works~\cite{lalonde2007learning,zhu2015learning}, we rank synthesized images from each dataset by the model's prediction, and visualize images in the 0\textsuperscript{th}, 25\textsuperscript{th}, 50\textsuperscript{th}, 75\textsuperscript{th}, 100\textsuperscript{th} percentile of the ``fakeness'' score from our model's output.

In most datasets, we observe little noticeable correlation between the model predictions and the visual quality of the synthesized images.
However, there is a weak correlation in the BigGAN and StarGAN datasets; qualitative examples are shown in Figure~\ref{fig:ranking}. 
As the ``fakeness'' scores are higher, the images tend to contain more visible artifacts which deteriorate the visual quality. 
This implies that our model might learn to capture perceptual realism under this task.
However, since the correlation is not observed in other datasets, it is more likely that the model learns features more towards low-level CNN artifacts. \supp{Examples across all datasets are provided in Appendix~\ref{sec:rank_appendix}.}

\myparagraph{Artifacts of CNN image synthesis}
Inspired by Zhang~\etal~\cite{zhang2019detecting}, we visualize the average frequency spectra from each dataset to study the artifacts generated by CNNs, as shown in Figure~\ref{fig:spectrums}. 
Following prior work, we perform a simple form of high-pass filtering (subtracting the image from its median blurred version) before calculating the Fourier transform, as it provides a more informative visualization~\cite{marra2019gans}. 
For each dataset, we average over 2000 randomly chosen images (or the entire set, if it is smaller).

We note that there are many interesting patterns visible in these visualizations.
While the real image spectra generally look alike (with minor variations due to differences in the datasets), there are distinct patterns visible in images generated by different CNN models. 
Furthermore, the repeated period patterns in these spectra may be consistent with aliasing artifacts, a cue considered by~\cite{zhang2019detecting}. 
Interestingly, the most effective unconditional GANs (BigGAN, ProGAN) contain relatively few such artifacts. Also, DeepFake images does not contain obvious artifacts. We note that DeepFake images have gone through various pre- and post-processing, where the synthesized face region is resized, blended, and compressed with MPEG. These operations perturbs the low-level image statistic, which may cause the frequency patterns to not emerge with this visualization method.
\section{Discussion}

Despite the alarm that has been raised by the rapidly improving quality of image synthesis methods, our results suggest that today's CNN-generated images retain \emph{detectable fingerprints} that distinguish them from real photos. This allows forensic classifiers to generalize from one model to another without extensive adaptation.

However, this does not mean that the current situation will persist.  Due to the difficulties in achieving Nash equilibria, none of the current GAN-based architectures are optimized to convergence, {\em i.e.} the generator never wins against the discriminator.
Were this to change, we would suddenly find ourselves in a situation when synthetic images are completely indistinguishable from real ones. 

Even with the current techniques, there remain practical reasons for concern.  
First, even the best forensics detector will have some trade-off between true detection and false-positive rates. Since a malicious user is typically looking to create a single fake image (rather than a distribution of fakes), they could simply hand-pick the fake image which happens to pass the detection threshold. 
Second, malicious use of fake imagery is likely be deployed on a social media platform (Facebook, Twitter, YouTube, \etc), so the data will undergo a number of often aggressive transformations (compression, resizing, re-sampling, \etc).
While we demonstrated robustness to some degree of JPEG compression, blurring, and resizing, much more work is needed to evaluate how well the current detectors can cope with these transformations \emph{in-the-wild}.
Finally, most documented instances of effective deployment of visual fakes to date have been using classic ``shallow'' methods, such as Photoshop.  We have experimented with running our detector on the face-aware liquify dataset from~\cite{wang2019detecting}, and found that our method performs at chance on this data.  This suggests that shallow methods exhibit fundamentally different behavior than deep methods, and should not be neglected.  

We note that detecting fake images is just one small piece of the puzzle of how to  combat the threat of visual disinformation.
Effective solutions will need to incorporate a wide range of strategies, from technical to social to legal. 
\myparagraph{Acknowledgements} {\small We'd like to thank Jaakko Lehtinen, Taesung Park, Jacob (Minyoung) Huh, \camready{Hany Farid, and Matthias Kirchner} for helpful discussions. We are grateful to Xu Zhang, \camready{Lakshmanan Nataraj, and Davide Cozzolino} for significant help with comparisons to~\cite{zhang2019detecting,nataraj2019detecting,cozzolino2018forensictransfer}, respectively. This work was funded, in part, by DARPA MediFor, Adobe gift and grant from the UC Berkeley Center for Long-Term Cybersecurity.
The views, opinions and/or findings expressed are those of the authors and should not be interpreted as representing the official views or policies of the Department of Defense or the U.S. Government.}

{\small
\bibliographystyle{ieee_fullname}
\bibliography{cnndet}

\begin{thebibliography}{10}\itemsep=-1pt

\bibitem{deepfakes2019faceswap}
Deepfakes faceswap github repository.
\newblock \url{https://github.com/ deepfakes/faceswap}.

\bibitem{faced}
Faced.
\newblock \url{https://github.com/iitzco/faced}.

\bibitem{deepfakes}
Faceswap.
\newblock \url{https://faceswap.dev/}.

\bibitem{whichfaceisreal}
Which face is real?
\newblock http://www.whichfaceisreal.com/.

\bibitem{agarwal2017photo}
Shruti Agarwal and Hany Farid.
\newblock Photo forensics from jpeg dimples.
\newblock In {\em 2017 IEEE Workshop on Information Forensics and Security
  (WIFS)}, 2017.

\bibitem{azulay2018deep}
Aharon Azulay and Yair Weiss.
\newblock Why do deep convolutional networks generalize so poorly to small
  image transformations?
\newblock {\em JMLR}, 2019.

\bibitem{bau2019seeing}
David Bau, Jun-Yan Zhu, Jonas Wulff, William Peebles, Hendrik Strobelt, Bolei
  Zhou, and Antonio Torralba.
\newblock Seeing what a gan cannot generate.
\newblock In {\em ICCV}, 2019.

\bibitem{bradski2008learning}
Gary Bradski and Adrian Kaehler.
\newblock {\em Learning OpenCV: Computer vision with the OpenCV library}.
\newblock "O'Reilly Media, Inc.", 2008.

\bibitem{brock2018large}
Andrew Brock, Jeff Donahue, and Karen Simonyan.
\newblock Large scale gan training for high fidelity natural image synthesis.
\newblock In {\em ICLR}, 2019.

\bibitem{chen2018learning}
Chen Chen, Qifeng Chen, Jia Xu, and Vladlen Koltun.
\newblock Learning to see in the dark.
\newblock In {\em CVPR}, 2018.

\bibitem{chen2017photographic}
Qifeng Chen and Vladlen Koltun.
\newblock Photographic image synthesis with cascaded refinement networks.
\newblock In {\em ICCV}, 2017.

\bibitem{choi2018stargan}
Yunjey Choi, Minje Choi, Munyoung Kim, Jung-Woo Ha, Sunghun Kim, and Jaegul
  Choo.
\newblock Stargan: Unified generative adversarial networks for multi-domain
  image-to-image translation.
\newblock In {\em CVPR}, 2018.

\bibitem{cozzolino2015splicebuster}
Davide Cozzolino, Giovanni Poggi, and Luisa Verdoliva.
\newblock Splicebuster: A new blind image splicing detector.
\newblock In {\em 2015 IEEE International Workshop on Information Forensics and
  Security (WIFS)}, 2015.

\bibitem{cozzolino2018forensictransfer}
Davide Cozzolino, Justus Thies, Andreas R{\"o}ssler, Christian Riess, Matthias
  Nie{\ss}ner, and Luisa Verdoliva.
\newblock Forensictransfer: Weakly-supervised domain adaptation for forgery
  detection.
\newblock {\em arXiv preprint arXiv:1812.02510}, 2018.

\bibitem{dai2019second}
Tao Dai, Jianrui Cai, Yongbing Zhang, Shu-Tao Xia, and Lei Zhang.
\newblock Second-order attention network for single image super-resolution.
\newblock In {\em CVPR}, 2019.

\bibitem{farid2016photo}
Hany Farid.
\newblock {\em Photo forensics}.
\newblock MIT Press, 2016.

\bibitem{he2016deep}
Kaiming He, Xiangyu Zhang, Shaoqing Ren, and Jian Sun.
\newblock Deep residual learning for image recognition.
\newblock In {\em CVPR}, 2016.

\bibitem{huh2018fighting}
Minyoung Huh, Andrew Liu, Andrew Owens, and Alexei~A Efros.
\newblock Fighting fake news: Image splice detection via learned
  self-consistency.
\newblock In {\em ECCV}, 2018.

\bibitem{isola2017image}
Phillip Isola, Jun-Yan Zhu, Tinghui Zhou, and Alexei~A Efros.
\newblock Image-to-image translation with conditional adversarial networks.
\newblock In {\em CVPR}, 2017.

\bibitem{johnson2016perceptual}
Justin Johnson, Alexandre Alahi, and Li Fei-Fei.
\newblock Perceptual losses for real-time style transfer and super-resolution.
\newblock In {\em ECCV}, 2016.

\bibitem{karras2018progressive}
Tero Karras, Timo Aila, Samuli Laine, and Jaakko Lehtinen.
\newblock Progressive growing of gans for improved quality, stability, and
  variation.
\newblock In {\em ICLR}, 2018.

\bibitem{karras2019style}
Tero Karras, Samuli Laine, and Timo Aila.
\newblock A style-based generator architecture for generative adversarial
  networks.
\newblock In {\em CVPR}, 2019.

\bibitem{karras2019analyzing}
Tero Karras, Samuli Laine, Miika Aittala, Janne Hellsten, Jaakko Lehtinen, and
  Timo Aila.
\newblock Analyzing and improving the image quality of stylegan.
\newblock In {\em CVPR}, 2020.

\bibitem{kingma2014adam}
Diederik~P Kingma and Jimmy Ba.
\newblock Adam: A method for stochastic optimization.
\newblock In {\em ICLR}, 2015.

\bibitem{lalonde2007learning}
Jean-Fran\c{c}ois Lalonde and Alexei~A. Efros.
\newblock Using color compatibility for assessing image realism.
\newblock In {\em ICCV}, 2007.

\bibitem{li2019diverse}
Ke Li, Tianhao Zhang, and Jitendra Malik.
\newblock Diverse image synthesis from semantic layouts via conditional imle.
\newblock In {\em ICCV}, 2019.

\bibitem{lin2014microsoft}
Tsung-Yi Lin, Michael Maire, Serge Belongie, James Hays, Pietro Perona, Deva
  Ramanan, Piotr Doll{\'a}r, and C~Lawrence Zitnick.
\newblock Microsoft coco: Common objects in context.
\newblock In {\em ECCV}, 2014.

\bibitem{liu2018large}
Ziwei Liu, Ping Luo, Xiaogang Wang, and Xiaoou Tang.
\newblock Large-scale celebfaces attributes (celeba) dataset.

\bibitem{marra2018detection}
Francesco Marra, Diego Gragnaniello, Davide Cozzolino, and Luisa Verdoliva.
\newblock Detection of gan-generated fake images over social networks.
\newblock In {\em 2018 IEEE Conference on Multimedia Information Processing and
  Retrieval (MIPR)}, 2018.

\bibitem{marra2019gans}
Francesco Marra, Diego Gragnaniello, Luisa Verdoliva, and Giovanni Poggi.
\newblock Do gans leave artificial fingerprints?
\newblock In {\em 2019 IEEE Conference on Multimedia Information Processing and
  Retrieval (MIPR)}, 2019.

\bibitem{nataraj2019detecting}
Lakshmanan Nataraj, Tajuddin~Manhar Mohammed, BS Manjunath, Shivkumar
  Chandrasekaran, Arjuna Flenner, Jawadul~H Bappy, and Amit~K Roy-Chowdhury.
\newblock Detecting gan generated fake images using co-occurrence matrices.
\newblock {\em Electronic Imaging}, 2019.

\bibitem{o2012exposing}
James~F O'Brien and Hany Farid.
\newblock Exposing photo manipulation with inconsistent reflections.
\newblock {\em ACM Trans. Graph.}, 2012.

\bibitem{odena2016deconvolution}
Augustus Odena, Vincent Dumoulin, and Chris Olah.
\newblock Deconvolution and checkerboard artifacts.
\newblock {\em Distill}, 2016.

\bibitem{park2019SPADE}
Taesung Park, Ming-Yu Liu, Ting-Chun Wang, and Jun-Yan Zhu.
\newblock Semantic image synthesis with spatially-adaptive normalization.
\newblock In {\em CVPR}, 2019.

\bibitem{perez2003poisson}
Patrick P{\'e}rez, Michel Gangnet, and Andrew Blake.
\newblock Poisson image editing.
\newblock {\em ACM Transactions on graphics (TOG)}, 2003.

\bibitem{platt1999probabilistic}
John Platt et~al.
\newblock Probabilistic outputs for support vector machines and comparisons to
  regularized likelihood methods.
\newblock 1999.

\bibitem{popescu2005exposing}
Alin~C Popescu and Hany Farid.
\newblock Exposing digital forgeries by detecting traces of resampling.
\newblock {\em IEEE Transactions on signal processing}, 2005.

\bibitem{rao2016deep}
Yuan Rao and Jiangqun Ni.
\newblock A deep learning approach to detection of splicing and copy-move
  forgeries in images.
\newblock In {\em 2016 IEEE International Workshop on Information Forensics and
  Security (WIFS)}, 2016.

\bibitem{rossler2019faceforensics++}
Andreas R{\"o}ssler, Davide Cozzolino, Luisa Verdoliva, Christian Riess, Justus
  Thies, and Matthias Nie{\ss}ner.
\newblock Face{F}orensics++: Learning to detect manipulated facial images.
\newblock In {\em ICCV}, 2019.

\bibitem{russakovsky2015imagenet}
Olga Russakovsky, Jia Deng, Hao Su, Jonathan Krause, Sanjeev Satheesh, Sean Ma,
  Zhiheng Huang, Andrej Karpathy, Aditya Khosla, Michael Bernstein, et~al.
\newblock Imagenet large scale visual recognition challenge.
\newblock {\em IJCV}, 2015.

\bibitem{torralba2011unbiased}
Antonio Torralba and Alexei~A Efros.
\newblock Unbiased look at dataset bias.
\newblock In {\em CVPR}, 2011.

\bibitem{ulyanov2018deep}
Dmitry Ulyanov, Andrea Vedaldi, and Victor Lempitsky.
\newblock Deep image prior.
\newblock In {\em CVPR}, 2018.

\bibitem{wang2019fakespotter}
Run Wang, Lei Ma, Felix Juefei-Xu, Xiaofei Xie, Jian Wang, and Yang Liu.
\newblock Fakespotter: A simple baseline for spotting ai-synthesized fake
  faces.
\newblock {\em arXiv preprint arXiv:1909.06122}, 2019.

\bibitem{wang2019detecting}
Sheng-Yu Wang, Oliver Wang, Andrew Owens, Richard Zhang, and Alexei~A Efros.
\newblock Detecting photoshopped faces by scripting photoshop.
\newblock In {\em ICCV}, 2019.

\bibitem{wang2018non}
Xiaolong Wang, Ross Girshick, Abhinav Gupta, and Kaiming He.
\newblock Non-local neural networks.
\newblock In {\em CVPR}, 2018.

\bibitem{yu2015lsun}
Fisher Yu, Ari Seff, Yinda Zhang, Shuran Song, Thomas Funkhouser, and Jianxiong
  Xiao.
\newblock Lsun: Construction of a large-scale image dataset using deep learning
  with humans in the loop.
\newblock {\em arXiv preprint arXiv:1506.03365}, 2015.

\bibitem{yu2019attributing}
Ning Yu, Larry Davis, and Mario Fritz.
\newblock Attributing fake images to gans: Analyzing fingerprints in generated
  images.
\newblock In {\em ICCV}, 2019.

\bibitem{zhang2018self}
Han Zhang, Ian Goodfellow, Dimitris Metaxas, and Augustus Odena.
\newblock Self-attention generative adversarial networks.
\newblock In {\em ICML}, 2019.

\bibitem{zhang2019making}
Richard Zhang.
\newblock Making convolutional networks shift-invariant again.
\newblock In {\em ICML}, 2019.

\bibitem{zhang2019detecting}
Xu Zhang, Svebor Karaman, and Shih-Fu Chang.
\newblock Detecting and simulating artifacts in gan fake images.
\newblock In {\em WIFS}, 2019.

\bibitem{zhou2016learning}
Bolei Zhou, Aditya Khosla, Agata Lapedriza, Aude Oliva, and Antonio Torralba.
\newblock Learning deep features for discriminative localization.
\newblock In {\em CVPR}, 2016.

\bibitem{zhou2018learning}
Peng Zhou, Xintong Han, Vlad~I Morariu, and Larry~S Davis.
\newblock Learning rich features for image manipulation detection.
\newblock In {\em CVPR}, 2018.

\bibitem{zhu2015learning}
Jun-Yan Zhu, Philipp Kr{\"a}henb{\"u}hl, Eli Shechtman, and Alexei~A. Efros.
\newblock Learning a discriminative model for the perception of realism in
  composite images.
\newblock In {\em ICCV}, 2015.

\bibitem{zhu2017unpaired}
Jun-Yan Zhu, Taesung Park, Phillip Isola, and Alexei~A Efros.
\newblock Unpaired image-to-image translation using cycle-consistent
  adversarial networks.
\newblock In {\em ICCV}, 2017.

\end{thebibliography}
}

\pagebreak
\setcounter{section}{0}
\renewcommand\thesection{\Alph{section}}
\renewcommand{\thefootnote}{\arabic{footnote}}

\noindent{\Large\bf Appendix}

\section{Additional Analysis}
\subsection{Additional ranking visualizations}
\label{sec:rank_appendix}
In %
we rank ordered the fake images according to how ``fake" the classifier deemed them to be. These full ranking results are included in the following link: \url{https://peterwang512.github.io/CNNDetection/ranking/}. We randomly select 20 real and 20 fake images from each dataset, and rank all images based on our \textbf{(Blur+JPEG (0.1))} model's scores. Note that there is a clear separation between real and fake images, where the real images have lower ``fakeness'' score and vice versa. Moreover, we observe the synthetic images ranked more ``real'' are super resolution (SAN) outputs, and the ones ranked more ``fake'' are CRN and IMLE outputs. However, we observe little noticeable correlation
between the model predictions and the visual quality of the
synthesized images in each dataset, where BigGAN and StarGAN images are the exceptions.

\subsection{Effect of dataset size}
\label{sec:dsize_appendix}

\begin{table*}[h]
  {\small
    \centering
    \resizebox{1.\linewidth}{!}{
    \begin{tabular}{ccccccc ccccccccccc c}
    \toprule

    \multirow{3}{*}{Family} & \multirow{3}{*}{Name} & \multicolumn{5}{c}{Training settings} & \multicolumn{11}{c}{Individual test generators} & Total \\
    \cmidrule(lr){3-7} \cmidrule(lr){8-18} \cmidrule(lr){19-19}

    & & \multirow{2}{*}{Train} & \multirow{2}{*}{Input} & \multirow{2}{*}{\shortstack[c]{No.\\Class}} & \multicolumn{2}{c}{Augments} & \multirow{2}{*}{\shortstack[c]{Pro-\\GAN}} & \multirow{2}{*}{\shortstack[c]{Style-\\GAN}} & \multirow{2}{*}{\shortstack[c]{Big-\\GAN}} & \multirow{2}{*}{\shortstack[c]{Cycle-\\GAN}} & \multirow{2}{*}{\shortstack[c]{Star-\\GAN}} & \multirow{2}{*}{\shortstack[c]{Gau-\\GAN}} & \multirow{2}{*}{CRN} & \multirow{2}{*}{IMLE} & \multirow{2}{*}{SITD} & \multirow{2}{*}{SAN} & \multirow{2}{*}{\shortstack[c]{Deep-\\Fake}} &
     \multirow{2}{*}{\bf mAP}\\ 
    \cmidrule(lr){6-7}
    & & & & & Blur & JPEG & \\
    \midrule

Nataraj~\etal~\cite{nataraj2019detecting} & -- & CycleGAN & Co-occur. mtx & -- & & & 76.4 & 96.5 & 56.4 & \gray{100.} & 88.2 & 56.2 & 58.7 & 83.1 & 39.6 & 46.1 & 55.1 & 68.8 \\
Cozzolino~\etal~\cite{cozzolino2018forensictransfer} & ForensicTransfer & ProGAN & HF residual & -- & & & \gray{88.9} & 77.9 & 79.5 & 77.2 & 91.7 & 83.3 & {\bf 99.9} & 31.3 & 72.8 & 90.8 & 79.2 & 79.3 \\
\midrule

& DIP & ProGAN-DIP & RGB & -- & \checkmark & \checkmark & 62.0 & 52.3 & 61.7 & 62.4 & {\bf 100.} & 49.0 & 98.2 & 38.6 & 92.8 & {\bf 93.1} & 63.1 & 70.3 \\

& Blur+JPEG (Big) & BigGAN & RGB & 1000 & \checkmark & \checkmark & {\bf 85.1} & 82.4 & \gray{100.} & 86.2 & 87.4 & 96.7 & 79.7 & 82.6 & 91.2 & 71.9 & 60.3 & 83.9 \\

\cdashline{2-19}

\multirow{9}{*}{Ours} & 2-class & ProGAN & RGB & 2 & \checkmark & \checkmark & \gray{98.8} & 78.3 & 66.4 & 88.7 & 87.3 & 87.4 & 94.0 & 97.3 & 85.2 & 52.9 & 58.1 & 81.3 \\
& 4-class & ProGAN & RGB & 4 & \checkmark & \checkmark & \gray{99.8} & 87.0 & 74.0 & 93.2 & 92.3 & 94.1 & 95.8 & 97.5 & 87.8 & 58.5 & 59.6 & 85.4 \\
& 8-class & ProGAN & RGB & 8 & \checkmark & \checkmark & \gray{99.9} & 94.2 & 78.9 & 94.3 & 91.9 & 95.4 & 98.9 & 99.4 & 91.2 & 58.6 & 63.8 & 87.9 \\
& 16-class & ProGAN & RGB & 16 & \checkmark & \checkmark & \gray{100.} & 98.2 & 87.7 & 96.4 & 95.5 & {\bf 98.1} & 99.0 & {\bf 99.7} & {\bf 95.3} & 63.1 & {\bf 71.9} & {\bf 91.4} \\

\cdashline{2-19}
 & 10\% data & ProGAN & RGB & 20 & \checkmark & \checkmark & \gray{100.} & 93.2 & 82.3 & 94.1 & 93.2 & 97.1 & 96.8 & 99.4 & 88.2 & 58.1 & 63.5 & 87.8 \\
& 20\% data & ProGAN & RGB & 20 & \checkmark & \checkmark & \gray{100.} & 96.8 & 85.9 & 95.9 & 93.6 & 97.9 & 98.7 & 99.5 & 90.2 & 61.8 & 65.2 & 89.6 \\
& 40\% data & ProGAN & RGB & 20 & \checkmark & \checkmark & \gray{100.} & 97.8 & 87.5 & 96.0 & 95.3 & 98.1 & 98.2 & 99.3 & 91.2 & 61.4 & 67.9 & 90.2 \\
& 80\% data & ProGAN & RGB & 20 & \checkmark & \checkmark & \gray{100.} & 98.1 & 88.1 & 96.4 & 95.4 & 98.0 & 98.9 & 99.4 & 93.0 & 63.8 & 65.1 & 90.6 \\

\cdashline{2-19}
& Blur+JPEG (0.5) & ProGAN & RGB & 20 & \checkmark & \checkmark & \gray{100.} & {\bf 98.5} & {\bf 88.2} & {\bf 96.8} & 95.4 & {\bf 98.1} & 98.9 & 99.5 & 92.7 & 63.9 & 66.3 & 90.8 \\

    \bottomrule
    \end{tabular}}
    }
    \vspace{-1.5mm}
    \caption{\camready{{\bf Additional evaluations.} We evaluate other baseline models, classifiers trained with DIP and BigGAN images, respectively, and classifiers trained with various dataset size. Same as Table 2 in the main text, we show the average precision (AP) of the models tested across 11 generators. For comparison, we include the ablations on the number of classes and the {\bf Blur+JPEG (0.5)} model's results, which are presented in the main text. Symbols $\checkmark$ means the augmentation is applied with $50\%$ or probability at training. The color coding scheme is identical to that of Table 2 in the main text. We note that when only the dataset size is reduced, AP dropped less comparing to reducing the number of classes. Also, the model trained with ProGAN outperforms the baselines, \textbf{DIP} and \textbf{Blur+JPEG (Big).} %
    }}
    \label{tab:sup_dsize}
\end{table*} We include additional ablation studies on the effect of dataset size, and the results are shown in Table~\ref{tab:sup_dsize}. To compare with the dataset diversity ablation in Section 4.3 of the main text, we train 4 additional models with 10\%, 20\%, 40\%, 80\% of the entire dataset respectively, while having all 20 LSUN classes included in the training set. \camready{Same augmentation scheme as \textbf{Blur+JPEG (0.5)} is applied to all models.} We observe much less reduction in generalization performance, indicating data diversity, comparing to dataset size, contributes more towards better CNN detection in general.

\subsection{Comparison to training on a different model}
\label{sec:biggan_appendix}
\camready{To evaluate the choice of training architecture, we also include a model that is trained \textit{solely} on BigGAN. To prepare the training data, we generate 400k fake images from an ImageNet-pretrained $256\times256$ BigGAN model~\cite{brock2018large}, and take 400k ImageNet images with the same class distribution as real images. For comparison, we train the model with the same data augmentation as \textbf{Blur+JPEG (0.5)}. We denote this model as \textbf{Blur+JPEG (Big)}.
We see in Table~\ref{tab:sup_dsize} that this model also exhibits generalization, albeit with slightly lower results in most cases. 
One explanation for this is that while our ProGANs model was trained on an ensemble (one model per class), BigGAN images were generated with a single model.}

\subsection{Training with images generated with a deep image prior}
\label{sec:dip_appendix}
\camready{Instead of generating fake images with GANs, which have limited representational capacity and hence large synthesis errors, we consider an ``oracle" generation method based on the deep image prior (DIP)~\cite{ulyanov2018deep}. We ask what the {\em very best} reconstruction of an image is achievable via a given network architecture, regardless of the synthesis task. For each synthesized image in our dataset, we train a {\em different} network to reconstruct it by minimizing $\ell_1$ loss:
\begin{equation}
    \min_\theta ||f(\theta_i) - I_i||_1,
\end{equation}
where $f(\theta_i)$ is the image generated by a neural network parameterized with weights $\theta_i$ and $I_i$ is a real image. We use the reconstructed image $f(\theta_i)$ as an instance of a fake image. During reconstruction, we use the Adam optimizer~\cite{kingma2014adam} with $\beta_1 = 0.9$, $\beta_2 = 0.999$, and optimized with a decreasing learning rate: $0.01\rightarrow 0.001 \rightarrow 0.0001$. For each learning rate we optimize for 2000 iterations.
}

\camready{As training data, we take 44k real images randomly sampled from ImageNet~\cite{russakovsky2015imagenet}, and ``fake'' images are the reconstruction by the generator of ProGAN (and hence 44k different networks). We take DIP images optimized for 1000, 2000, 3000, 4000, 5000, 6000 iterations into  our ``fake'' image set. We then train a classifier on this dataset, and we over-sample the real images by 6 times to balance the classes. All training configurations and augmentations are same as \textbf{Blur+JPEG (0.5)}. This model is denoted as \textbf{DIP} in Tab.~\ref{tab:sup_dsize}.}

\camready{We note although this model does not perform as well as the model directly trained on ProGAN images, but it is able to detect several datasets, including StarGAN, CRN, SITD, and SAN. This indicates that low-level artifacts shares across different methods, but just leveraging on those may not be sufficient for a general detection.}

\subsection{Comparison to other baselines}
\label{sec:baseline_appendix}
\camready{In the main text, we compared with Zhang~\etal~\cite{zhang2019detecting}, a state-of-the-art in GAN detection, and outperform it across different synthesis methods.
In addition, we include the performance of Nataraj~\etal~\cite{nataraj2019detecting}, another GAN detection method trained on co-occurrence matrices of images, and Cozzolino~\etal\cite{cozzolino2018forensictransfer}, a few-shot single-target domain adaption method trained on HF filtered images. For Cozzolino~\etal, we evaluate the ProGAN/CycleGAN model. Both methods are evaluated on $256\times256$ images in a zero-shot setting, and if the image is larger than 256 pixels, it is center-cropped to 256 pixels. The results are in Tab.~\ref{tab:sup_dsize}.}

\subsection{Other evaluation metrics}
\label{sec:twoshot}

To help clarify the threshold-less AP evaluation metric, we also computed several other metrics (\tbl{tab:twoshotresults}). We provide the precision and recall curve on each dataset from our {\bf(Blur+JPEG (0.1))} model in Figure~\ref{fig:pr_curve}. We give the {\em uncalibrated} generalization accuracy of the model on the test distribution, by simply using the classifier threshold we learned during training, and {\em oracle} accuracy that chooses the threshold that maximizes accuracy on the test set. We also consider a {\em two-shot} regime where we have access to one real and one fake image from each dataset, and only the model's \textit{threshold} is adjusted during the two-shot calibration process. 

We calibrate the model by a single random real and fake pair, and we augment the image pair by taking $224\times224$ random crops 128 times. The images are passed into the model to get the logits, which are then fitted by a logistic regression (the method is also known as Platt scaling~\cite{platt1999probabilistic}). We take the bias learned from the logistic regression to adjust the base rate of our model. Specifically, we apply the bias to our model's logit and then take the sigmoid to get the calibrated probability.

\begin{figure}
    \centering
    \includegraphics[width=1.\linewidth]{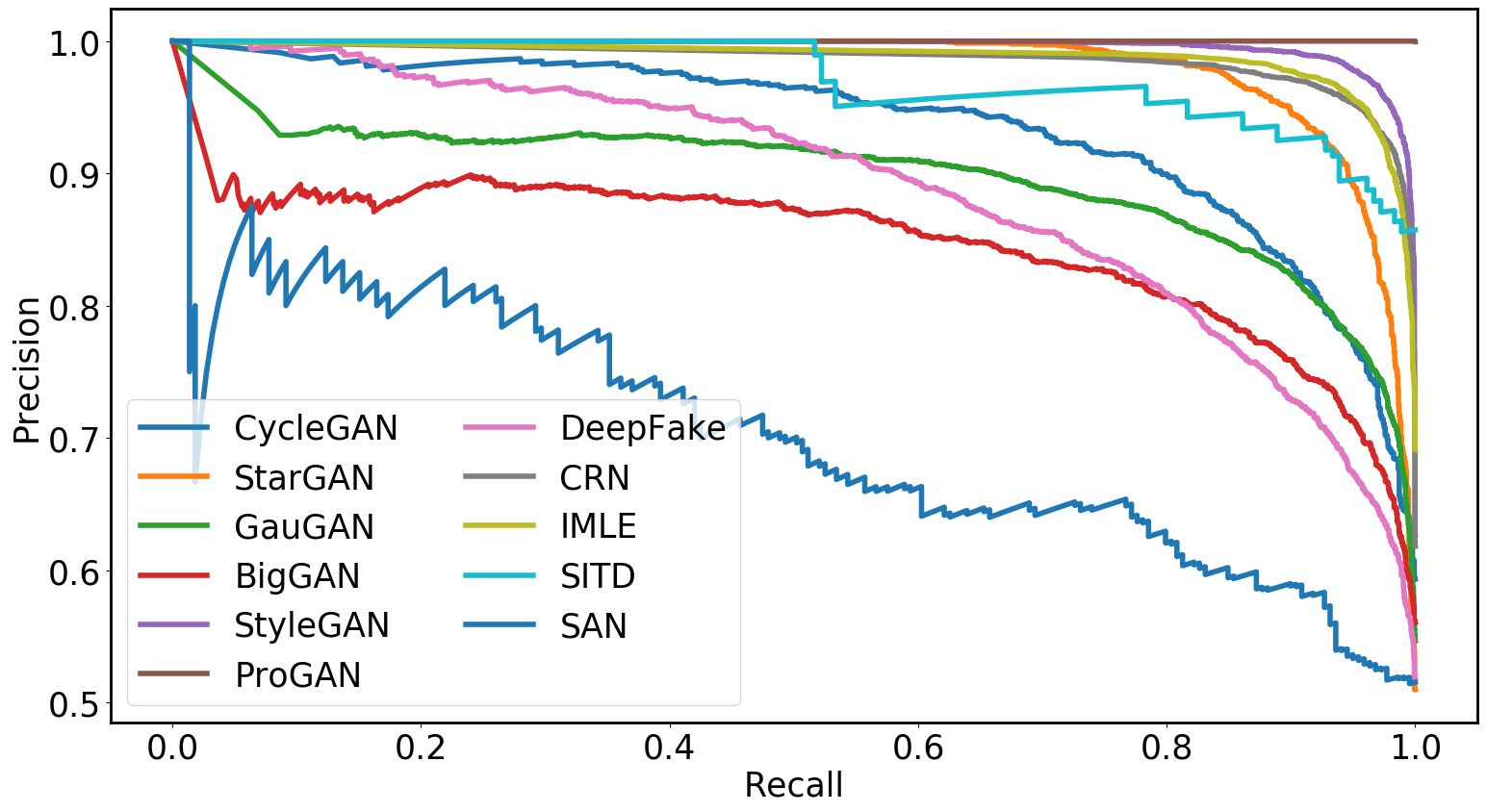}
  \caption{{\bf Precision and recall curves.} The PR curves on each dataset from the \textbf{(Blur+JPEG (0.1))} model are shown. Note that AP is defined as the area under the PR curve. Higher AP indicates better trade-off between precision and recall, and vice versa.}
  \label{fig:pr_curve}
\end{figure}
\begin{table*}[h]
    \centering
    \resizebox{1.\linewidth}{!}{
    \begin{tabular}{@{}lcccccccccc@{}}
    \toprule
    & StyleGAN & BigGAN & CycleGAN & StarGAN & GauGAN & CRN & IMLE & SITD & SAN & DeepFake \\
    \midrule 
    Uncalibrated & 87.1 & 70.2 & 85.2 & 91.7 & 78.9 & 86.3 & 86.2 & 90.3 & 50.5 & 53.5 \\
    Oracle & 96.8 & 81.1 & 86.3 & 92.8 & 85.5 & 95.3 & 95.4 & 92.8 & 68.0 & 80.7 \\
    \cdashline{1-11}
    Two-shot & 91.9 & 74.0& 82.4& 86.0 & 79.1 & 91.6& 91.2 & 88.7 & 54.8 & 65.7 \\
    \bottomrule
    \end{tabular}
    }
    \caption{\textbf{Two-shot classifier calibration.} We show the accuracy of the classifiers directly trained from ProGAN (``uncalibrated''), after calibrating the threshold given two examples in the test distribution (``two-shot'') and an upper bound, given a perfect calibration (``oracle'').}
    \label{tab:twoshotresults}
\end{table*}

\begin{table*}[h]
    \centering
    \resizebox{1.\linewidth}{!}{
    \begin{tabular}{@{}ccccccccccccccc@{}}
    \toprule
    Horse & Zebra & Summer & Winter & Apple & Orange & Facades & Cityscape & Map & Ukiyoe & Vangogh & Cezanne & Monet & Photo & \textbf{Avg.} \\
    \midrule 
    62.1 & 87.5 & 83.2 & 88.0 & 90.5 & 87.7 & 100. & 66.6 & 78.0 & 85.4 & 76.9 & 82.8 & 56.2 & 86.8 & 80.8 \\

    \bottomrule
    \end{tabular}
    }
    \caption{\camready{\textbf{CycleGAN testcase.} We evaluate the uncalibrated accuracy of \textbf{Blur+JPEG (0.1)} model tested on each CycleGAN category. We note that our model is still able to perform well above chance (50\%) even if not directly trained on any CycleGAN images.}}
    \label{tab:cyclegantest}
\end{table*}

\subsection{Detecting GAN images from the internet}
\camready{Unfortunately, there are currently no collections of ``in-the-wild'' CNN-generated image datasets which we can evaluate with our model. As a ``proxy'' testcase, we scraped 1k real face and 1k fake faces from \url{whichfaceisreal.com}~\cite{whichfaceisreal}. This is a website containing StyleGAN-generated faces and real faces in 1024 pixels, with all images compressed into JPEG. We tested our \textbf{Blur+JPEG (0.1)} model on this testset in two scenarios: (1) directly center crop images to 224 pixels without resizing (matching how we test StyleGAN) or (2) resize to 256 pixels then center crop to 224 pixels. Without resizing, the model gets 83.6\% accuracy and 93.2\% AP. With resizing, the model drops to 74.9\% accuracy and 82.6\% AP, still well above chance (50\%). This indicates our model can be robust to resizing and in-the-wild JPEG compression. However, maintaining similar performance after significant post-processing (e.g., heavy resizing) remains challenging.}

\subsection{CycleGAN testcase}
\camready{While prior works on GAN detection~\cite{marra2018detection,nataraj2019detecting,zhang2019detecting} train on CycleGAN images and evaluate generalization across CycleGAN \emph{categories}, our method is not trained on any CycleGAN images and tests generalization across \emph{methods} (a significantly harder task).
Nonetheless, we still observe comparable performance in terms of AP (Tab. 2 in the main text) when compared to Zhang~\etal~\cite{zhang2019detecting}. %
For a further comparison, we include our \textbf{Blur+JPEG (0.1)} model's accuracy on each CycleGAN category in Tab.~\ref{tab:cyclegantest}.}

\section{Implementation Details}
\subsection{Dataset Collection}
\label{sec:data_appendix}
\paragraph{ProGAN~\cite{karras2018progressive} \protect\footnote{\tiny \url{https://github.com/tkarras/progressive_growing_of_gans}}} We take 20 officially released ProGAN models pretrained on LSUN~\cite{yu2015lsun} airplane, bicycle, bird, boat, bottle, bus, car, cat, chair, cow, dining table, dog, horse, motorbike, person, potted plant, sheep, sofa, train, tv-monitor respectively. Following the official code, we sample the synthetic images with $z \sim N(0, I)$, and generate real images by center cropping the images just on the long edge (center crop length is exactly the length of the short edge) and then resizing to $256\times256$ \\

\paragraph{StyleGAN~\cite{karras2019style} \protect\footnote{\tiny \url{https://github.com/NVlabs/stylegan}}} We take officially released StyleGAN models pretrained on LSUN~\cite{yu2015lsun} bedroom, cat and car, with size $256\times256$, $256\times256$ and $512\times384$ respectively. We download the released synthesized images, all of which are generated with 0.5 truncation, and following the code, we generate real images by resizing to the according size of each category.\\

\paragraph{StyleGAN2~\cite{karras2019analyzing} \protect\footnote{\tiny \url{https://github.com/NVlabs/stylegan2}}} We take officially released StyleGAN2 config-F models pretrained on LSUN~\cite{yu2015lsun} church, cat, horse and car, with size $256\times256$, $256\times256$, $256\times256$ and $512\times384$ respectively. We download the released synthesized images, all of which are generated with 0.5 truncation, and following the code, we generate real images by resizing to the according size of each category.\\

\paragraph{BigGAN~\cite{brock2018large} \protect\footnote{\tiny \url{https://tfhub.dev/s?q=biggan}}} We take officially released BigGAN-deep model pretrained on $256\times256$ ImageNet images. Following the official code, we sample the images with uniform class distribution and with 0.4 truncation; also, we generate real images by center cropping the images just on the long edge (center crop length is exactly the length of the short edge) and then resizing to $256\times256$.

\paragraph{CycleGAN~\cite{zhu2017unpaired} \protect\footnote{\tiny \url{https://github.com/junyanz/pytorch-CycleGAN-and-pix2pix}}} We take officially released CycleGAN models: apple2orange, orange2apple, horse2zebra, zebra2horse, summer2winter, winter2summer, and generate real and fake image pairs out of all six categories. Pre-processed real images and synthetic images are directly generated from the released code.

\paragraph{StarGAN~\cite{choi2018stargan} \protect\footnote{\tiny \url{https://github.com/yunjey/stargan}}} We take officially released StarGAN model pretrained on CelebA~\cite{liu2018large}, and generate real and fake image pairs. Pre-processed real images and synthetic images are directly generated from the released code.

\paragraph{GauGAN~\cite{park2019SPADE} \protect\footnote{\tiny \url{https://github.com/NVlabs/SPADE}}} We take officially released GauGAN model pretrained on COCO~\cite{lin2014microsoft}, and generate real and fake image pairs. Pre-processed real images and synthetic images are directly generated from the released code.

\paragraph{CRN~\cite{chen2017photographic} \protect\footnote{\tiny \url{https://github.com/CQFIO/PhotographicImageSynthesis}}} We take officially released CRN model pretrained on GTA, and generate synthesized images from pre-processed segmentation maps. Pre-processed real images and segmentation maps are downloaded from the IMLE repository.

\paragraph{IMLE~\cite{li2019diverse} \protect\footnote{\tiny \url{https://github.com/zth667/Diverse-Image-Synthesis-from-Semantic-Layout}}} We take officially released IMLE model pretrained on GTA, and generate synthesized images from pre-processed segmentation maps. Pre-processed real images and segmentation maps are downloaded from the official repository.

\paragraph{SITD~\cite{chen2018learning} \protect\footnote{\tiny \url{https://github.com/cchen156/Learning-to-See-in-the-Dark}}} We take officially released pretrained model and the dataset by Sony and Fuji cameras from the repository. Pre-processed real images and synthetic images are directly generated from the released code.

\paragraph{SAN~\cite{dai2019second} \protect\footnote{\tiny \url{https://github.com/daitao/SAN}}} We take both the ground truth and the officially released 4x super-resolution predictions on the standard benchmark datasets: Set5, Set14, BSD100 and Urban100. The synthetic images are directly downloaded from the repository.

\paragraph{DeepFake~\cite{rossler2019faceforensics++} \protect\footnote{\tiny \url{https://github.com/ondyari/FaceForensics}}} We download raw manipulated and original image sequences in the validation and test split of the Deepfakes dataset. We extracted all frames from the videos, and in each frame a face is detected and cropped using Faced~\cite{faced}. Similar to \cite{rossler2019faceforensics++}, our dataset is comprised entirely of cropped faces.

\subsection{Training details}
\label{sec:traindetail_appendix}
To train the classifiers, we use the Adam optimizer~\cite{kingma2014adam} with $\beta_1 = 0.9$, $\beta_2 = 0.999$, batch size 64, and initial learning rate $10^{-4}$. Learning rate are dropped by $10\times$ if after 5 epochs the validation accuracy does not increase by $0.1\%$, and we terminate training at learning rate $10^{-6}$. One exception is that, in order to balance training iterations with the size of the training set, for the \{2, 4, 8, 16\}-class models and the \{10, 20, 40, 80\}\%-data models, the learning rate is dropped if the validation accuracy plateaus for \{50, 25, 13, 7\} epochs instead.

\end{document}